\documentclass[times, review, 10pt]{elsarticle}

\usepackage[hyphens]{url}  
\usepackage{amssymb}
\usepackage{amsmath}
\usepackage{algorithm}
\usepackage{algpseudocode}
\usepackage{amsmath}
\usepackage{bbding}
\usepackage{booktabs}
\usepackage{colortbl}
\usepackage{subcaption}
\usepackage{tcolorbox}
\usepackage{tablefootnote}
\usepackage[hidelinks]{hyperref}
\usepackage{newfloat}
\usepackage{listings}
\usepackage[normalem]{ulem}

\journal{Pattern Recognition}

\begin{document}

\begin{frontmatter}



\title{GeoNav: Empowering MLLMs with Dual-Scale Geospatial Reasoning for Language-Goal Aerial Navigation} 


\author[label1,label2]{Haotian Xu} 
\author[label1,label2]{Yue Hu\corref{cor1}}
\author[label3]{Chen Gao}
\author[label1,label2]{Zhengqiu Zhu}
\author[label1,label2]{Yong Zhao}
\author[label1,label2]{Quanjun Yin}
\affiliation[label1]{organization={College of Systems Engineering, National University of Defense Technology},
            city={Changsha},
            postcode={410073}, 
            country={China}}
\affiliation[label2]{organization={State Key Laboratory of Digital Intelligent Modeling and Simulation},
            city={Changsha},
            postcode={410073}, 
            country={China}}
\affiliation[label3]{organization={BNRist, Tsinghua University},
            city={Beijing},
            postcode={100084}, 
            country={China}}
\cortext[cor1]{Corresponding author. Email: huyue11@nudt.edu.cn}
\begin{abstract}
Language-goal aerial navigation requires UAVs to localize targets in the complex outdoors such as urban blocks based on textual instruction. The indoor methods are often hard to scale to urban scenes due to ambiguous objects, limited visual field and spatial reasoning.
In this work, we propose \textbf{GeoNav}, a multi-modal agent for long-range aerial navigation with geospatial awareness. GeoNav operates in three phases—landmark navigation, target search, and precise localization—mimicking human coarse-to-fine spatial reasoning patterns. To support such reasoning, it dynamically builds dual-scale spatial representations. The first is a global but schematic cognitive map, which fuses prior geographic knowledge and embodied visual cues into a top-down and explicit annotated form. It enables fast navigation to the landmark region via intuitive map-based reasoning. The second is a local but delicate scene graph representing hierarchical spatial relationships between landmarks and objects, utilized for accurate target localization. On top of the structured memory, GeoNav employs a spatial chain-of-thought mechanism to enable MLLMs with efficient and interpretable action-making across stages.
On the CityNav benchmark, GeoNav surpasses the current SOTA up to 18.4$\%$ in success rate and significantly eliminate navigation error. The ablation studies highlight the importance of each module, position structured spatial perception as the key to advanced UAV navigation.
\end{abstract}




\end{frontmatter}



\section{Introduction}
\label{sec:intro}

Language-goal aerial navigation \citep{DBLP:conf/iccv/LiuZQ0ZW23, jiang2025uevavd} is an emerging task in embodied intelligence research, where an unmanned aerial vehicle (UAV) is required to comprehend natural language instructions and interpret spatial concept to navigate to a target in unseen outdoor environments. It holds significant potential for urban governance and public services, such as emergency response, aerial logistics, and security patrols \citep{11127476, wandelt2023aerial, doschl2024say, cai2025neusis}. 

Urban environments are inherently rich in points of interest (POIs), i.e., landmarks in navigation contexts, and their geographic information are easily accessed. However, most existing aerial navigation methods disregard to use such knowledge. Instead, they adopt an egocentric vision-language matching paradigm \citep{gao2025openfly}, akin to architecture used in ground-level navigation \citep{he2023frequency, shah2023lm, schumann2024velma}. While effective for small-scale settings, these methods usually fail when scaled to larger urban environments.

Comparatively, this paper focus on a specific setting with practical meanings, i.e., aerial navigation based on landmark geographic priors, a brand-new and challenging task which has recently substantiated by the benchmark CityNav \citep{lee2025citynav}. Although UAVs are equipped the ability to observe from elevated viewpoints and explore the expansive aerial space, the task presents three critical challenges.
\begin{figure}
\centering
\includegraphics[width=\linewidth]{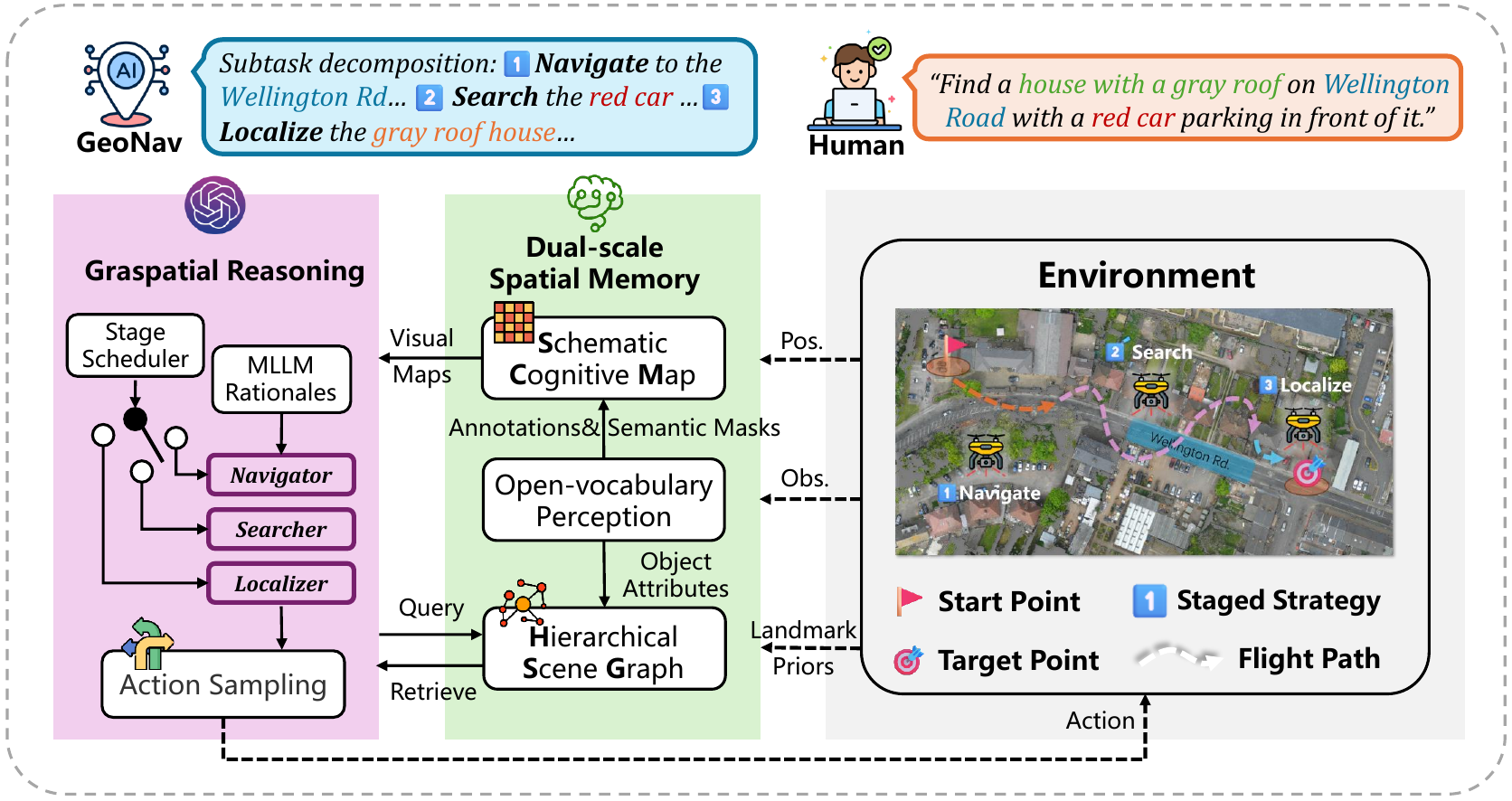}
\caption{Illustration of the GeoNav agent structure. SCM is dynamically built and used for reasoning in navigation and search, while HSG is retrieved for target localization.}
\label{fig:framework}
\end{figure}

\begin{itemize}
\item[$\bullet$] \textbf{Ambiguous Visual Semantics:} Unlike indoor scenarios, the urban entities considerable visually ambiguity~\citep{sarkar2024gomaa, zhou2025urbench}. \textit{The prevalence of structurally similar objects  (e.g., a specific building in a complex, a vehicle in a parking lot) necessitates identification based on geospatial referring expression.}

\item[$\bullet$] \textbf{Multi-scale Spatial Reasoning:} Urban environments contain macro-scale elements beyond immediate visibility \citep{zheng2023spatial} (e.g., road networks, districts) and directly observable micro-scale entities \citep{rottensteiner2014results} (e.g., buildings, vehicles). \textit{The absence of an effective spatial representation 
integrating multi-level knowledge poses fundament al challenges for large-scale and adptive spatial reasoning.}

\item[$\bullet$] \textbf{Long-horizon Planning:} A long-range aerial navigation task involves multiple phases, including when the target object is out of view \citep{zhao2025aerial, jones2023path, HE2024110511}.
\textit{The lack of context-aware decision-making renders the agent with inflexible strategies for changing situations.
}
\end{itemize}

To address the challenges, we propose \textbf{GeoNav}, a zero-shot agentic approach that bridges urban geographic knowledge and embodied perception for aerial navigation. As depicted in Figure \ref{fig:framework}, GeoNav leverages landmark priors and multi-modal large language models (MLLMs) to perform explicit spatial reasoning through structured memory and task decomposition.

GeoNav operates under a \textbf{three-stage and coarse-to-fine workflow}, inspired by how humans localize targets in unfamiliar large-scale cities: (1) Landmark Navigation — navigating toward coarse-grained geographic landmarks with known locations and even geometric appearances (e.g., \textit{``near the train station''}), (2) Target Search — searching within the local landmark region for objects matching the goal description, and (3) Precise Localization — final decision making to pinpoint the target location. By integrating multi-resolution spatial knowledge, the staged process allows for more efficient search than end-to-end methods.

To facilitate the navigation toward the specified landmark, GeoNav builds a global but \textbf{schematic cognitive map (SCM)}, explicitly annotating the locations of the UAV, landmarks and observed objects. The sketchy but intuitive map makes GeoNav aware of the desired direction and distance.
Then, to support reasoning for localizing the target, GeoNav locally constructs a \textbf{hierarchical scene graph (HSG)} as structured spatial memory around the landmark. It contains two kind of nodes: (1) Geographic Nodes, such as blocks or landmarks, constructed from prior knowledge, and (2) Object Nodes with semantic and spatial attributes extracted from top-down observations. The relationships between nodes (e.g., \textit{“near”},\textit{“inside”}) are dynamically updated as the exploration.

To connect perception and reasoning, we design an LLM-based task planner for strategy adjusting and a \textbf{Stage-conditioned Reasoning} mechanism. At each stage of navigation, the system generates stage-specific rationale prompts including subgoal and desired state descriptions that query the MLLM with structured spatial memory to produce action-level advice.

There are three main contributions in this work.
\begin{itemize}
    \item[$\bullet$] We propose a stage-aware scheduling and reasoning mechanism for language-goal aerial navigation, wherein the task is decomposed and tackled progressively via structured and multi-modal rationales with chain of thoughts (CoTs).
    \item[$\bullet$] To support cross-scale navigation, GeoNav fuses textual geographic priors, instructions and visual observations to construct dual-scale resolutions of spatial representations to facilitate fast navigation and accurate localization.
    \item[$\bullet$] Extensive evaluation on CityNav, a challenging urban navigation benchmark, demonstrates the substantial improvement of GeoNav over typical baselines in navigational success rate and efficiency.
\end{itemize}

\section{Related Work}
We discuss the related studies from the application domain and core techniques: aerial vision-and-language navigation and multi-modal LLM-driven spatial reasoning.
\subsection{Aerial Vision-and-Language Navigation}
Aerial Vision-and-Language Navigation (VLN) requires UAVs to navigate complex environments via natural language instruction and embodied perception. Early Aerial VLN methods proposed foundational benchmarks \citep{DBLP:conf/acl/FanCJZZW23, zhao-etal-2025-cityeqa} and simulators \citep{tian2025uavs, DBLP:conf/iccv/LiuZQ0ZW23}, while later studies refined spatial representations. For instance, STMR \citep{gao2024aerial} employs an LLM-driven agent guided by hybrid semantic-topo-metric maps, and NavAgent \citep{DBLP:journals/corr/abs-2411-08579} aligns multiscale visual cues with rich textual descriptions, significantly enhancing navigation in outdoor settings.

Recent strides in dataset construction also propels this field. OpenUAV \citep{wang2024towards} offers diverse scenes and realistic controls, while AeroVerse \citep{DBLP:journals/corr/abs-2408-15511} unifies simulation, pretraining, and evaluation across various tasks. CityNav \citep{lee2025citynav} further enriches the landscape with 3D point clouds and urban 2D maps, highlighting the effectiveness of human demonstrations over shortest-path baselines.
Despite these advances, geographic aerial navigation in urban-scale settings remains challenging. While CityNav provides relevant data, existing models struggle due to insufficient spatial reasoning capabilities. This work aims to address this on the geographic aerial navigation tasks, which are more aligned with practical UAV application scenarios.

\subsection{MLLM-driven Spatial Reasoning}

Spatial reasoning is an important approach to achieving embodied AI. Unlike text-based spatial reasoning, visual thinking does not rely on spatial relationship templates, but instead extracts richer spatial information from natural images and videos \citep{cai2025spatialbot, DBLP:conf/coling/LeiYCLL25, DBLP:journals/corr/abs-2310-11441, zhao-etal-2025-urbanvideo}. Research shows that MLLMs have the ability to mimic how humans construct spatial representations \citep{yang2025thinking, zhan2025open3dvqa}. For example, MVOT \citep{pmlr-v267-li25cz} achieves visual reasoning through a multi-modal generative thinking process. MAM \citep{he2024memory} adaptively reduces the visual and text noise to enhance the navigation memory. TopV-Nav \citep{DBLP:journals/corr/abs-2411-16425} and MapNav \citep{zhang-etal-2025-mapnav} incorporate annotation text into semantic map representations to enhance the spatial memory of large models.

Unlike STMR, GeoNav adopts a visual cognitive map (with editable geospatial annotations) into the chain of thought. Compared to TopV-Nav and MapNav, two indoor navigation methods based on MLLM, this paper considers a more realistic urban task setting involving larges space and objects that are harder to distinguish.

\section{Task Formulation}
\label{sec:formatting}
We follow the task of language-goal aerial navigation with geographic information \citep{lee2025citynav}, where the input is defined as a triplet $\mathcal{I} = \{ \mathcal{T} , \mathcal{K}, \mathcal{O}_t \mid t\in[0,T)\}$. A natural language description $\mathcal{T}$, composed of a series of words $\{ w_1, w_2, \dots, w_N \}$ specifies the navigation target along with surrounding landmarks and their spatial relationships. 
At timestep $t = 0$, the agent is provided with a set of geographic priors $\mathcal{K} = \{ k_1, \dots, k_m \}$, where each $k_i = \langle l_i, c_i\rangle$ represents a landmark description $l_i \in \mathcal{L}$ and a region contour $c_i \in \mathcal{C}$ respectively. Here, $c_i$ is an ordered sequence of coordinate points $\mathbf{p}_i^j \in \mathbb{R}^2$, i.e., $c_i = (\mathbf{p}_i^1, \dots, \mathbf{p}_i^n)$.
At timestep $1\leq t<T$, the current observation $\mathcal{O}_t$ consists of an egocentric top-down RGB image $I_t$ from its onboard camera and its current pose $\xi_t=(x_t,y_t,z_t,\theta_t)$. $(x_t,y_t)$ denotes the GPS coordinate of the UAV in the horizontal plane, $z_t$ represents its altitude, and $\theta_t$ indicates its heading direction.

The agent selects an action $\mathcal{A}_t$ from a discrete action space comprising eight planar movements (East, West, South, North, Northeast, Northwest, Southeast and Southwest, 5 meters per step), two turning actions (turn left/right, 30 degrees each time), two vertical movements (ascending and descending, 5 meters per step), and stop. An episode is considered successful if the UAV stops within 200 steps and is within 20 meters of the target location. In our setting, the MLLM reasoning is invoked every 10 steps.

\begin{figure}
  \centering
  \includegraphics[width=\textwidth]{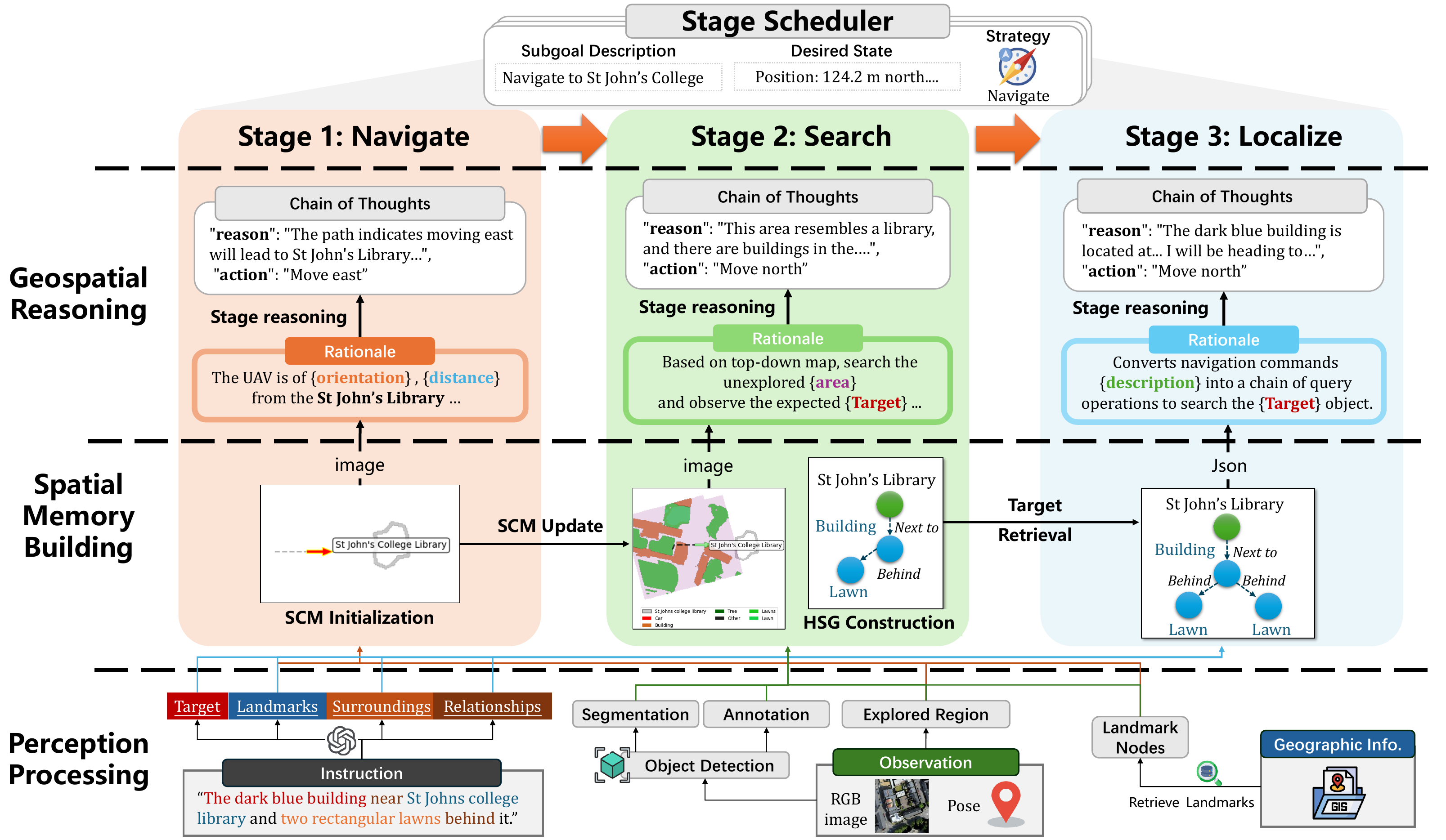}
  \caption{Illustration of the GeoNav workflow, featured by coarse-to-fine stages and perception-memory-reasoning processes.}
  \label{fig:pipeline}
\end{figure}

\section{Method}
GeoNav comprises three key modules: a multi-stage navigation strategy (MNS), a schematic cognitive map (SCM), and a hierarchical scene graph (HSG). As illustrated in Figure \ref{fig:pipeline}, the workflow is presented along two dimensions. \textbf{Horizontally}, it progresses through coarse-to-fine stages, including navigate, search, and localize, with subgoals defined by a stage scheduler. The SCM is updated during the stages of navigation and search, while HSG is constructed when searching, but is used for eventual localization. \textbf{Vertically}, each time the MLLM is invoked, GeoNav operates across three hierarchical layers: \textbf{(1) Perception Processing (Bottom).} Extracts multi-modal features from instructions and visual observations to ground landmarks and objects. \textbf{(2) Spatial Memory Building (Middle).} Maintains a persistent world memory, evolving from a global SCM for navigation to a fine-grained HSG for localization. \textbf{(3) Geospatial Reasoning (Top).} Engages an MLLM in ``Chain of Thoughts'' reasoning to generate action plans based on current memory and rationale.

\subsection{Stage-aware Reasoning}
This subsection details how MNS prompts an MLLM to make decisions. We first introduce the stage scheduler to decompose a task into subtasks with different stages. Navigation and search stages adopt the vision-language-action (VLA) paradigm, receiving a fused spatial cognitive map as input and generating actions. In contrast, the localize stage uses the HSG constructed in the search stage, queries the possible target, and controls the drone to reach the location with deterministic actions.

\subsubsection{Stage Scheduler}
At the task beginning, the stage scheduler employs an MLLM to act as a high-level planner. The planner mimics human problem-solving by decomposing long-horizon tasks into manageable sub-goals.
As detailed in line 1 of Algorithm \ref{alg:geonav_main_simple}, given a natural language instruction $\mathcal{T}$ and a set of geographic priors $\mathcal{K}$, the planner function $\Pi$ generates a structured navigation plan $\mathrm{P}$:
    \begin{equation}
\mathrm{P} = \Pi (\mathcal{T},\mathcal{K}),
\end{equation}
where $\mathrm{P}$ is an ordered list of sub-goals. In line 5, $sg_i$ is defined as a triplet containing the goal description $g_i$, the desired state $s_i$, and the execution strategy $\sigma_i$. The planner selects the strategy $\sigma_i$ from a predefined set of operational modes:
\begin{equation}
    \Sigma = \{ \text{Navigate}, \text{Search}, \text{Localize} \}.
\end{equation}
The decomposition rules are clarified in the \ref{appendix_stage}.

\subsubsection{Multi-modal Reasoning}
As shown in line 6 to line 16 of Algorithm \ref{alg:geonav_main_simple}, the agent activates the corresponding strategy to achieve sub-goals. The geospatial reasoning for each stage is as follows:
\begin{itemize}
    \item \textbf{Navigate} is used for long-range landmark positioning. It primarily uses the global SCM to guide the UAV efficiently toward the landmark region specified in the sub-goal. At this stage, the rationale $\Phi_i$ (orange box in Figure \ref{fig:pipeline}) is composed of the distance $d_i$ and angle $\phi_i$ between the current pose $\xi_t$ of the agent and the centroid ${\bar c_i}$ of the landmark contour. 
    \item \textbf{Search} focuses on curiosity-driven local exploration \citep{pathak2017curiosity} based on the updating SCM to observe potential targets. Once in proximity to the landmark, it dynamically constructs and updates the HSG. At this stage, the rationale $\Phi_i$ (the green box) is the tips for region exploration, including the current position $\xi_t$ and the visual field $[(x_{min},y_{min}),(x_{max},y_{max})]$.
    \item \textbf{Localize} is activated for the target localization. It involves querying the rich spatial relationships within the completed HSG to pinpoint the exact target node. At this stage, the rationale (the blue box) follows a template process that translates the query into a structured chain of operations over the graph.
\end{itemize}

At each step, the MLLM makes a decision guided by the active sub-goal $g_i$, a corresponding prompt rationale $\Phi_i$ and the corresponding spatial representation $\mathcal{R}_t$. The model outputs textual reasoning process $\Gamma_t$ and the next action $\mathcal{A}_t$:
\begin{equation}
    \langle \Gamma_t,\mathcal{A}_t \rangle= MLLM(\mathcal{R}_t, \Phi_i, g_i, s_i \mid \Sigma).
\end{equation}
For more details of the Chain-of-Thought prompts used, please refer to \ref{appendix_cot}.

\begin{algorithm}[thp]
\caption{Multi-stage Navigation Process}
\label{alg:geonav_main_simple}
\begin{algorithmic}[1]
\Require Instruction $\mathcal{T}$, Geographic Priors $\mathcal{K}$
\Ensure Action sequence $\mathcal{A}_{0:T}$
\State \textbf{Planning:} $\mathrm{P} = \{sg_1, sg_2, \dots\} \leftarrow \text{StageScheduler}(\mathcal{T}, \mathcal{K})$ \Comment{Stage Scheduler}
\State \textbf{Initialization:} SCM $M_0 \leftarrow \text{SCM\_{Init}}(\mathcal{K})$, HSG $G_0 \leftarrow \emptyset$, Step $t \leftarrow 0$, $i \leftarrow 1$
\While{not \textit{Done}}
    \State Obtain $\mathcal{O}_t= \langle I_t, \xi_t \rangle$ and update SCM $M_t \leftarrow \text{SCM\_Update}(\mathcal{O}_t, M_t)$
    \State Get current subgoal $sg_{i} = \langle g_{i}, s_{i}, \sigma_{i} \rangle$
    
    \If{$\sigma_{i} == \text{Navigate}$}
        \State $\mathcal{A}_t \leftarrow \text{MLLM}(M_t, g_{i})$ \Comment{Generate action based on global view $M_t$}
        
    \ElsIf{$\sigma_{i} == \text{Search}$}
        \State Incrementally update HSG: $G_t \leftarrow \text{HSG\_Update}(G_{t-1}, M_t, \mathcal{T})$
        \State $\mathcal{A}_t \leftarrow \text{MLLM}(M_t, g_{i})$ \Comment{Generate action based on exploration}
        
    \ElsIf{$\sigma_{i} == \text{Localize}$}
         \State Target $v^* \leftarrow \text{HSG\_Query}(G_t, \mathcal{T})$
         \State $\mathcal{A}_t \leftarrow \text{DeterministicControl}(\xi_t, v^*)$ \Comment{Generate action based on controller}
    \EndIf
    \State Execute action $\mathcal{A}_t$
    \If{$\text{CheckCondition}(s_{i})$ is True} \Comment{Check status}
       \State $i \leftarrow i + 1$ \Comment{Switch to next stage}
    \EndIf
\EndWhile
\end{algorithmic}
\end{algorithm}

\subsection{Schematic Cognitive Map}
For effective long-range navigation, the agent requires a holistic model of the urban environment. Given the challenge that LLM/MLLM face in interpreting raw numerical coordinates \citep{merrill2023parallelism}, we introduce the SCM, denoted as $M$, which converts abstract geographic priors and real-time embodied observations into a unified, top-down visual representation that an MLLM can readily comprehend.

The construction of the map $M$ is an iterative process based on the agent's observations over time:
\begin{itemize}
    \item \textbf{Initialization ($M_0$)}: The initial map is generated by projecting the landmark contours $\{ c_i\}$ from the geographic priors $\mathcal{K}$ onto a 2D canvas.
    \item \textbf{Update ($M_t$)}: At each subsequent timestep $t>0$, the map is updated with perceptual information from the current observation $\mathcal{O}_t$. The update rule is defined as:
    \begin{equation}
M_t=M_{t-1} \cup \Delta_{\text {perceive}}\left(I_t\right).
\end{equation}
\end{itemize}
Here, $\Delta_{\text {perceive}}$ represents a perception function that processes top-down image $I_t$ from the UAV. It employs GroundDino for object detection and Segment Anything \citep{liu2024grounding, kirillov2023segment} for segmentation to extract semantic objects and map them to their corresponding world coordinates within $M_t$.

The resulting map $M_t$ integrates landmark and object layers with the trajectory of the UAV, forming a rich semantic representation that contains both static landmarks and dynamic objects, as illustrated in Figure \ref{fig:pipeline}. This map serves as the core visual input for the agent during the landmark navigation and target search stages. The detailed coordinate transformations for this process are described in \ref{appendix_SCM}.

\begin{figure}
\centering
\includegraphics[width=0.95\linewidth]{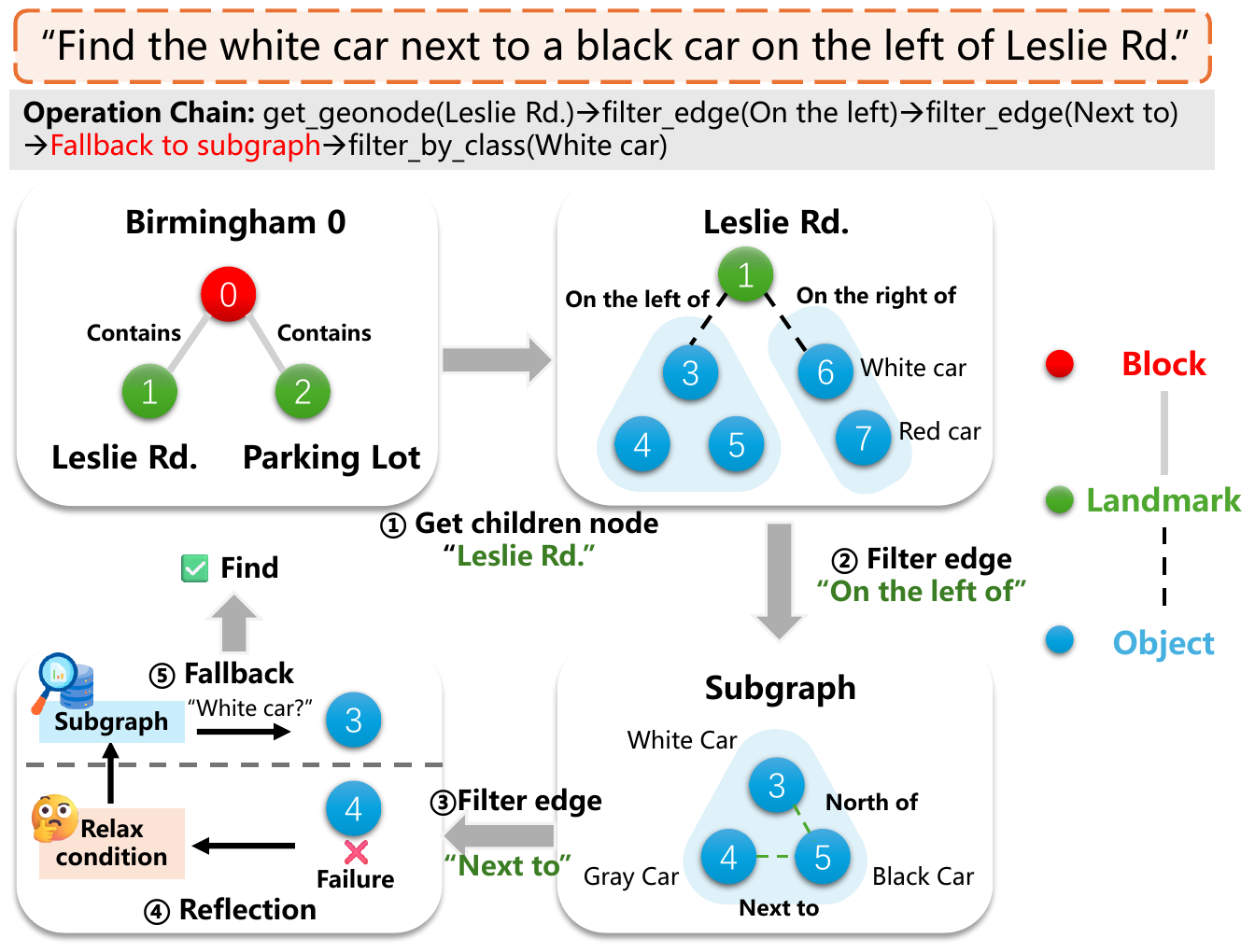}
\caption{An example on querying the hierarchical scene graph.}
\label{fig:sg}
\end{figure}

\subsection{Hierarchical Scene Graph}
In this subsection, we introduce the construction of our novel scene graph structure and describe the maintenance and retrieval for query a specific target.

\subsubsection{Scene Graph Structure}
The HSG, denoted as $G=(V,E)$ is composed of nodes $V$ and edges $E$. To model the urban scene at different granularities, as shown in Figure~\ref{fig:sg}, we define three kinds of hierarchical nodes. A block node represents a large area that contains multiple landmarks. Landmark nodes, i.e. $\mathcal{K}$ representing geographic POIs. Object nodes are detected from the UAV camera. Each object node $v_i \in V$ stores attributes including real-world positions $(x_i,y_i)$ and the semantic features. Edges connect nodes both within and across levels, representing rich spatial relationships like containment, adjacency, and relative direction.

\subsubsection{Scene Graph Construction}
The HSG is built and updated dynamically during the ``Search'' stage, once the agent has navigated to the landmark vicinity. As shown in the bottom left of Figure \ref{fig:pipeline}, visual grounding models detect the interested objects based on the parsed label from the given $\mathcal{T}$. Then, these objects in the camera view are segmented and unified into real-world coordinates to ensure a spatial reference for all nodes in the HSG.

In addition to the connections between blocks and landmarks, GeoNav introduces two other critical edge types: \textbf{Landmark-Object Edges} and \textbf{Object-Object Edges}.
For the former, the relationship $e(v_i, l_i)$ between an object $v_i$ and a landmark $l_i$ is determined by a rule-based function, $\varphi_{\text{rules}}$. This function is formally expressed as:
\begin{equation}
\label{eq:spatial_relation}
e(v_i, l_i) = \varphi_{\text{rules}}\Big( \|(x_i,y_i) - \bar c_i\| , \alpha(v_i, l_i) \Big),
\end{equation}
where $\alpha$ represents the relative angle, and the term $\|(x_i,y_i) - \bar c_i\|$ signifies the distance or displacement. The function $\varphi_{\text{rules}}$ maps these inputs to one element of a pre-defined spatial expression set, such as $\{ \text{contains}, \text{north\_of}, \text{northeast\_of}, \dots \}$.

To determine the spatial relationship \(e(v_i, v_j)\) between two observed objects, we leverage the reasoning capabilities of an MLLM, which processes the visual input $I_t$ to infer their relative layout (e.g., \textit{``next to''}, \textit{``behind''}):
\begin{equation}
\label{eq:spatial_relation_simple}
e(v_i, v_j) =\phi_{\text{MLLM}}(I).
\end{equation}

\subsubsection{Incremental Graph Update}
The HSG is incrementally updated at each timestep during the search phase. To avoid creating duplicate nodes for the same object observed from slightly different camera viewpoints, we merge new node candidates with existing nodes if they are sufficiently similar. This similarity is quantified by a comprehensive score $S_{ij}$ that fuses both spatial and semantic information:
\begin{equation}
\label{eq:similarity}
S_{ij} = (1-\gamma) \, S_{ij}^{\text{spatial}} + \gamma \, S_{ij}^{\text{semantic}},
\end{equation}
where $\gamma$ is a hyperparameter that balances the two components.

To ensure a mathematically coherent fusion, both components are formulated as bounded similarity scores. The spatial similarity $S_{ij}^{\text{spatial}}$ converts the Euclidean distance between two node positions, $p_i$ and $p_j$, into a score between 0 and 1 using a Gaussian kernel:
\begin{equation}
\label{eq:spatial_similarity}
S_{ij}^{\text{spatial}} = \exp(-\|p_i - p_j\|_2^2 / \sigma^2),
\end{equation}
where $\sigma$ is a scaling hyperparameter. This formulation ensures that closer nodes yield a higher similarity score.

The semantic similarity $S_{ij}^{\text{semantic}}$ is computed as the cosine similarity between the visual feature embeddings of the two nodes, which are extracted from our object detector. If the combined similarity score $S_{ij}$ exceeds a predefined threshold $\rho$, the nodes are considered to represent the same entity and are merged. The implementation details and hyperparameter values ($\gamma$, $\rho$, $\sigma$) are provided in \ref{appendix_HSG}.
\begin{algorithm}[htp]
\caption{HSG-based Target Retrieval}
\label{alg:hsg_query_detail}
\begin{algorithmic}[1]
\Require Instruction $\mathcal{T}$, HSG $G_t=(V,E)$, Current Pose $\xi_t$
\Ensure Target Position $\mathbf{p}^*$

\Function{HSG\_Query}{$\mathcal{T}, G_t, \xi_t$}
    \State \textbf{Step 1: Query Generation}
    \State $C \leftarrow \text{OpChain}(\mathcal{T})$ \Comment{Generate operation chain}
    \State $V_{cand} \leftarrow \text{ExecChain}(C, G_t)$ \Comment{Chain of queries on the graph}

    \State \textbf{Step 2: Selection or Fallback}
    \If{$V_{cand} \neq \emptyset$}
        \State $v^* \leftarrow \arg\max_{v \in V_{cand}} \text{Confidence}(v)$
    \Else \Comment{\textbf{Fallback Mechanism}}
        \For{$k \in \{1, \cdots ,max\_trial\}$}
            \State $C\_rel=\text{RelaxChain}(\mathcal{T})$
            \State $V_{rel} \leftarrow \text{ExecChain}(C\_rel, G_t)$
            \State $v^* \leftarrow \arg\max_{v \in V_{rel}} \text{Confidence}(v)$ \Comment{Find relaxed node}
        \EndFor
    \State \Return $v^*.\text{pos}$ \Comment{Directly localize target}
    \EndIf
\EndFunction
\end{algorithmic}
\end{algorithm}
\subsubsection{LLM-Driven Target Retrieval}
By reasoning on the HSG as a memory module, GeoNav can pinpoint a target with complex spatial relationships. We design an LLM-driven multi-hop query mechanism. As illustrated in Algorithm \ref{alg:hsg_query_detail}, this works as follows:
\begin{itemize}
    \item \textbf{Operation Chain Generation:} The MLLM converts the instruction into a chain of operations as presented in line 3 of Algorithm \ref{alg:hsg_query_detail}. The process is guided by a detailed prompt that provides a schema of available query operations and a few-shot demonstration.
    \item \textbf{Recursive Execution:} In line 4, the agent filters nodes and edges based on the operation chain to identify the target. The hierarchical structure is particularly effective for disambiguating between visually similar objects based on their relational context.
    \item \textbf{Failure Reflection:} If an initial query fails to accurately find the target, a fallback mechanism is triggered (from line 8 to 14). This system can automatically relax query constraints (e.g., changing a required \textit{``adjacent to''} relationship to the broader \textit{``contains''}) and recursively re-execute the query to ensure the most relevant target is found.
\end{itemize}
An example is provided in Figure \ref{fig:sg}. In this case, the target node is localized after fallback to query a semantic attribute \textit{``white car''}.
\section{Experiments}

\subsection{Experimental Settings}
\subsubsection{Dataset Preparation}
Our navigation dataset comes from the CityNav dataset \citep{lee2025citynav}, which provides detailed instructions of target, landmark, region, spatial relationships, etc. The average flight distance is 545 meters, which exceeds that of OpenUAV (255 meters), AVDN (144 meters), and OpenFly (less than 150 meters). 
Excluding the three main partitions—Validation Seen, Validation Unseen, and Test Unseen—we further divide the two unseen splits into three difficulty levels according to target distances.
The images are taken from SensatUrban \citep{hu2022sensaturban}, which collects orthographic projections of Birmingham (13 blocks) and Cambridge (33 blocks) to simulate RGB input obtained by UAVs. Meanwhile, the coordinates of landmarks are extracted from OSM maps and stored in the geographic information dataset CityRefer \citep{miyanishi2023cityrefer}.

\subsubsection{Evaluation Metrics}
To rigorously evaluate navigation performance, we adopt four typical metrics used in language-goal navigation \citep{wang2019reinforced, krantz2020beyond}. These standard metrics are: Navigation Error (NE), Success Rate (SR), Oracle Success Rate (OSR), and Success weighted by Path Length (SPL).
NE measures the Euclidean distance between the final position of the agent and the ground truth target. SR calculates the percentage of episodes in which the agent terminates within a predefined success threshold (20 meters). OSR evaluates whether the agent's trajectory ever approaches the target within the threshold, decoupling the grounding of the instruction from path execution. SPL is the success rate weighted by the ratio of the reference path length to the actual path length. The path length is calculated as the cumulative distance between consecutive navigation nodes.

\subsubsection{Implementation Details}
GeoNav employs GPT-4o \citep{hurst2024gpt} to process textual inputs to extract key elements such as objects and place names, analyzes images, and returns navigation plans. GroundingDINO \citep{liu2024grounding} serves as the foundation model for open-vocabulary object detection, while Segment Anything \citep{kirillov2023segment} is used for image segmentation. Further details can be found in \ref{appendix_implementation}.

\begin{table}[htp]
\centering
\scriptsize
\caption{Overall performance against the current SOTAs. Learning-based models are trained with human demonstrations (HD) and shortest path (SP) trajectories. All results are reported as mean ± standard deviation over 5 independent runs. \textbf{Bold} text indicates the best results under corresponding metrics other than human, while \underline{underlined} text marks the second best. The results of Seq2seq, CMA, and MGP are reproduced using the official implementation \tablefootnote{\label{citynavlink} available at: \protect\url{https://github.com/water-cookie/citynav}}.}

\label{tab:overall_results}
\setlength{\tabcolsep}{2.8pt}
\resizebox{\textwidth}{!}{
\begin{tabular}{@{}l c c c c c c c c c c c c c c c@{}}
\toprule
 & \multicolumn{3}{c}{Observation} & \multicolumn{4}{c}{Test unseen} & \multicolumn{4}{c}{Validation unseen} & \multicolumn{4}{c}{Validation seen}\\  
\cmidrule(lr){2-4} \cmidrule(lr){5-8} \cmidrule(lr){9-12} \cmidrule(lr){13-16}
\textbf{Method} & RGB & Instruction & Geo & NE$\downarrow$ & SR$\uparrow$ & OSR$\uparrow$ & SPL$\uparrow$  & NE$\downarrow$ & SR$\uparrow$ & OSR$\uparrow$ & SPL$\uparrow$ & NE$\downarrow$ & SR$\uparrow$ & OSR$\uparrow$ & SPL$\uparrow$ \\ 
\midrule 
RS      & -  & - & - &  208.8 ± 3.2  &  0.0 ± 0.0   &   1.4 ± 0.3  &  0.0 ± 0.0   &  223.0 ± 4.1  &  0.0 ± 0.0  &  1.4 ± 0.2   &  0.0 ± 0.1  &  222.3 ± 3.8  &  0.0 ± 0.0  &  0.9 ± 0.2   &  0.0 ± 0.0  \\
Greedy      & -  & \Checkmark & \Checkmark     &  105.6 ± 2.1  &  1.6 ± 0.4   &   3.2 ± 0.5  &  1.0 ± 0.3   &  103.0 ± 1.9  &  0.0 ± 0.1  &  4.9 ± 0.6   &  0.0 ± 0.0  &  78.8 ± 1.5  &  0.0 ± 0.1  &  2.7 ± 0.4   &  0.0 ± 0.0  \\
\midrule
Seq2Seq w/ SP  & \Checkmark  & \Checkmark & -  &  174.5 ± 3.8  &  1.7 ± 0.3   &   8.6 ± 0.7  &  1.7 ± 0.3   &  201.4 ± 4.2  &  1.1 ± 0.2  &  8.0 ± 0.6   &  1.0 ± 0.2  &  148.4 ± 2.9  &  4.5 ± 0.5  &  10.6 ± 0.8   &  4.5 ± 0.5 \\
Seq2Seq w/ HD  & \Checkmark  & \Checkmark & -  &  245.3 ± 5.1  &  1.5 ± 0.3   &   8.3 ± 0.7  &  1.3 ± 0.3   &  317.4 ± 6.3  &  0.8 ± 0.2  &  8.8 ± 0.7   &  0.6 ± 0.2  &  257.1 ± 4.8  &  1.8 ± 0.3  &  7.9 ± 0.6   &  1.6 ± 0.3 \\
CMA w/ SP  & \Checkmark  & \Checkmark & -      &  179.1 ± 3.9  &  1.6 ± 0.3   &   10.1 ± 0.8  &  1.6 ± 0.3   &  205.2 ± 4.3  &  1.1 ± 0.2  &  7.9 ± 0.6   &  1.1 ± 0.2  &  151.7 ± 3.1  &  3.7 ± 0.4  &  10.8 ± 0.8   &  3.7 ± 0.4  \\
CMA w/ HD  & \Checkmark  & \Checkmark & -      &  252.6 ± 5.3  &  0.8 ± 0.2   &   9.7 ± 0.8  &  0.8 ± 0.1   &  268.8 ± 5.2  &  0.7 ± 0.1  &  7.9 ± 0.6   &  0.6 ± 0.2  &  240.8 ± 4.6  &  1.0 ± 0.3  &  9.4 ± 0.7   &  0.9 ± 0.1  \\
MGP w/ SP & \Checkmark  & \Checkmark & \Checkmark  &  110.0 ± 2.3   &  4.5 ± 0.5   &  17.0 ± 1.2  &  4.3 ± 0.5   &  95.6 ± 2.1   &  3.6 ± 0.4  &  14.5 ± 1.0  &  3.5 ± 0.4  &  76.0 ± 1.8   &  5.6 ± 0.6   &  22.2 ± 1.4  &  5.4 ± 0.6  \\
MGP w/ HD & \Checkmark  & \Checkmark & \Checkmark  &  94.4 ± 2.0   &  5.6 ± 0.6   &  25.8 ± 1.6  &  \underline{5.3 ± 0.5}   &  75.3 ± 1.7   &  5.0 ± 0.5  &  21.9 ± 1.3  &  4.7 ± 0.5  &  \underline{60.9 ± 1.4}   &  \underline{7.2 ± 0.7}   &  \underline{34.8 ± 1.8}  &  \underline{6.6 ± 0.6}  \\
\midrule
NavGPT & \Checkmark  & \Checkmark & \Checkmark  &  342.9 ± 6.5   &  0.0 ± 0.2  &  2.4 ± 0.4  &  0.0 ± 0.3  &  345.5 ± 6.8   &  0.0 ± 0.0  &  4.5 ± 0.6  &  0.0 ± 0.0  &  234.5 ± 4.5   &  0.0 ± 0.1   &   4.3 ± 0.5  &  0.0 ± 0.0   \\
QwenVL-Max & \Checkmark  & \Checkmark & \Checkmark  &  455.4 ± 8.2   &  0.0 ± 0.0   &  1.2 ± 0.3  &  0.0 ± 0.0  &  324.2 ± 6.2   &  0.0 ± 0.0  &  2.8 ± 0.4  &  0.0 ± 0.0  &  573 ± 10.5  &   0.0 ± 0.0   &   0.9 ± 0.2  &  0.0 ± 0.0   \\
\midrule
TopV-Nav w/ MNS & \Checkmark  & \Checkmark & \Checkmark &  \underline{79.5 ± 1.7}   &  5.8 ± 0.6  &  35.9 ± 1.9  &  3.3 ± 0.4  &  77.4 ± 1.6   &  7.7 ± 0.7   &   24.5 ± 1.4  &  4.1 ± 0.5  &  102.7 ± 2.2   &  5.0 ± 0.5  &  34.2 ± 1.7  &  3.2 ± 0.4  \\
STMR w/ MNS & \Checkmark  & \Checkmark & \Checkmark   &  84.3 ± 1.8   &  \underline{7.5 ± 0.7}  &  \textbf{42.5 ± 2.1}  &  3.6 ± 0.4  &  \underline{73.0 ± 1.6}   &  \underline{8.2 ± 0.7}   &   \underline{25.1 ± 1.4}  &  \underline{5.9 ± 0.6}  &  103.8 ± 2.3   &  5.8 ± 0.6  &  29.1 ± 1.5  &  4.0 ± 0.5  \\
\midrule 
\rowcolor{gray!10} 
\textbf{GeoNav} & \Checkmark  & \Checkmark & \Checkmark  &  \textbf{73.5 ± 1.2}   &  \textbf{25.9 ± 1.1}  &  \underline{41.6 ± 1.8}  &  \textbf{16.0 ± 0.9}   &  \textbf{64.1 ± 1.1}   &  \textbf{16.9 ± 0.9}  &  \textbf{35.5 ± 1.5}  &  \textbf{9.9 ± 0.6}  &  \textbf{58.6 ± 1.0}   &  \textbf{21.8 ± 1.0}  &  \textbf{39.9 ± 1.6}  &  \textbf{12.3 ± 0.7}  \\
\midrule 
\rowcolor{gray!20} 
Human & \Checkmark  & \Checkmark & \Checkmark  &  9.8 ± 0.8   &  87.9 ± 2.3  &  95.3 ± 1.5  &  57.0 ± 2.1  &  9.4 ± 0.7   &  88.4 ± 2.2  &  95.5 ± 1.4  &  62.7 ± 2.3  &  9.1 ± 0.7   &  89.3 ± 2.1   &   96.4 ± 1.3  & 60.2 ± 2.2  \\
\bottomrule
\end{tabular}}
\end{table}

\subsection{Overall Performance}
\subsubsection{Baseline Methods}
We introduce three categories of methods for comparison with GeoNav. 
\begin{itemize}
\item[$\bullet$] \textbf{Rule-based Methods} includes the RS method, which selects actions randomly from the permissible actions. The greedy method navigates toward the landmark centroid and directly locates the target at that point.
\item[$\bullet$] \textbf{Learning-based Methods} includes mainstream aerial VLN methods Seq2Seq and CMA \citep{DBLP:conf/acl/FanCJZZW23}, as well as the SOTA method, MGP \citep{lee2025citynav}. 
\item[$\bullet$] \textbf{Zero-shot Methods}. We compare four zero-shot navigation methods, including the ego-centric view method NavGPT \citep{zhou2024navgpt} and native Qwen-VL-Max \citep{bai2025qwen2}. TopV-Nav \citep{DBLP:journals/corr/abs-2411-16425} and STMR \citep{gao2024aerial} based on a top-down map for comparison. For fairness, we add the MNS module to TopV-Nav and STMR to reach the landmarks fast and provide these four methods with prior landmark descriptions.
\end{itemize}

\subsubsection{Performance Analyses}
Table \ref{tab:overall_results} presents the overall performance of the primary methods on the CityNav dataset. \textit{Among them, GeoNav outperforms other methods across all four metrics.} Across the three splits, its SR surpasses the second-best method by 18.4\%, 8.7\%, and 14.6\%, respectively, while its SPL leads by 10.7\%, 4.0\%, and 5.7\%. In the test unseen, TopV-Nav and STMR achieve results close to those of GeoNav. In the validation unseen, STMR demonstrates comparable performance. In the validation seen, MGP significantly outperforms others, which can be attributed to its generalization on training. Furthermore, we note the performance gap between the current methods and humans. For instance, in the test unseen, the NE from human is only 9.8m, while SR reaches as high as 87.9\%. This phenomenon reveals the research challenge for current algorithmic designs to address this task.

\begin{table}[htp] 
\centering
\scriptsize
\caption{Generalization performance across difficulty levels of the test unseen split. All results are reported as mean ± standard deviation over 5 independent runs. \textbf{Bold} text indicates the best results under corresponding metrics, while \underline{underlined} text marks the second best.}
\label{tab:diffculty_results}
\setlength{\tabcolsep}{2.8pt}
\resizebox{\textwidth}{!}{
\begin{tabular}{@{}l c c c c c c c c c c c c c c c@{}}
\toprule
 & \multicolumn{3}{c}{Observation} & \multicolumn{4}{c}{Easy} & \multicolumn{4}{c}{Medium} & \multicolumn{4}{c}{Hard}\\  
\cmidrule(lr){2-4} \cmidrule(lr){5-8} \cmidrule(lr){9-12} \cmidrule(lr){13-16}
\textbf{Method} & RGB & Instruction & Geo & NE$\downarrow$ & SR$\uparrow$ & OSR$\uparrow$ & SPL$\uparrow$  & NE$\downarrow$ & SR$\uparrow$ & OSR$\uparrow$ & SPL$\uparrow$ & NE$\downarrow$ & SR$\uparrow$ & OSR$\uparrow$ & SPL$\uparrow$ \\ 
\midrule 
RS      & -  & - & - &  165.2 ± 2.8 &  0.0 ± 0.0  &   0.0 ± 0.0 &  0.0 ± 0.0   &  248.6 ± 4.5  &  0.0 ± 0.0  &  0.0 ± 0.0   &  0.0 ± 0.0  &  453.3 ± 8.2  &  0.0 ± 0.0  &  0.0 ± 0.0   &  0.0 ± 0.0  \\
Greedy      & -  & \Checkmark & \Checkmark     &  105.6 ± 2.1  &  1.6 ± 0.4   &   3.2 ± 0.5  &  1.0 ± 0.3   &  103.0 ± 1.9  &  0.0 ± 0.0  &  4.9 ± 0.6   &  0.0 ± 0.0  &  78.8 ± 1.5  &  0.0 ± 0.0  &  2.7 ± 0.4   &  0.0 ± 0.0  \\
\midrule
Seq2Seq w/ HD  & \Checkmark  & \Checkmark & -  &  122.8 ± 2.3  &  0.0 ± 0.0   &   6.3 ± 0.6  &  0.0 ± 0.0   &  189.6 ± 3.5  &  0.0 ± 0.0  &  0.0 ± 0.0   &  0.0 ± 0.0  &  276.5 ± 5.1  &  0.0 ± 0.0  &  0.0 ± 0.0   &  0.0 ± 0.0  \\
CMA w/ HD  & \Checkmark  & \Checkmark & -      &  118.3 ± 2.2  &  0.0 ± 0.0   &   10.4 ± 0.9  &  0.0 ± 0.0   &  193.8 ± 3.6  &  0.0 ± 0.0  &  0.0 ± 0.0   &  0.0 ± 0.0  &  315.2 ± 5.8  &  0.0 ± 0.0  &  0.0 ± 0.0   &  0.0 ± 0.0  \\
MGP w/ HD & \Checkmark  & \Checkmark & \Checkmark  &  81.8 ± 1.7   &  \underline{5.1 ± 0.5}   &  13.1 ± 1.1  &  \underline{4.7 ± 0.5}   &  \underline{69.4 ± 1.4}   &  \underline{9.1 ± 0.8}  &  13.1 ± 1.1  &  \underline{8.7 ± 0.8}  &  \underline{78.2 ± 1.6}   &  2.0 ± 0.3   &  3.4 ± 0.4  &  1.9 ± 0.3  \\
\midrule
NavGPT & \Checkmark  & \Checkmark & \Checkmark  &  301.5 ± 5.6   &  0.0 ± 0.0  &  7.5 ± 0.7  &  0.0 ± 0.0  &  327.3 ± 6.1   &  0.0 ± 0.0  &  6.3 ± 0.6  &  0.0 ± 0.0  &  401.6 ± 7.4   &  0.0 ± 0.0   &   0.0 ± 0.0  &  0.0 ± 0.0   \\
QwenVL-Max & \Checkmark  & \Checkmark & \Checkmark  &  454.5 ± 8.3   &  0.0 ± 0.0   &  4.8 ± 0.5  &  0.0 ± 0.0  &  450.8 ± 8.2   &  0.0 ± 0.0  &  4.2 ± 0.5  &  0.0 ± 0.0  &  465.5 ± 8.5  &   0.0 ± 0.0   &   0.0 ± 0.0  &  0.0 ± 0.0   \\
\midrule
TopV-Nav w/ MNS & \Checkmark  & \Checkmark & \Checkmark  &  \underline{75.2 ± 1.6}  &  4.8 ± 0.5 &  \textbf{61.9 ± 2.5} &  2.2 ± 0.3 &  89.2 ± 1.9  &  8.3 ± 0.8 &  \underline{27.1 ± 1.4} &  5.6 ± 0.6 &  152.8 ± 3.0  &  \underline{2.1 ± 0.3}  &   \underline{4.2 ± 0.5} &  \underline{2.0 ± 0.3} \\
STMR w/ MNS & \Checkmark  & \Checkmark & \Checkmark  &  80.8 ± 1.7  &  4.8 ± 0.5 &  52.4 ± 2.2 &  1.7 ± 0.3 &  87.3 ± 1.8  &  10.4 ± 0.9 &  25.0 ± 1.4 &  7.6 ± 0.7 &  151.1 ± 2.9  & 2.1 ± 0.3  &  2.1 ± 0.3 & 1.1 ± 0.2  \\
\midrule 
\textbf{GeoNav} & \Checkmark  & \Checkmark & \Checkmark  & \textbf{66.1 ± 1.4}  & \textbf{26.1 ± 1.1}  & \underline{40.1 ± 1.7} & \textbf{15.2 ± 0.9}  & \textbf{53.8 ± 1.2}  & \textbf{22.9 ± 1.0} & \textbf{39.6 ± 1.7} & \textbf{17.1 ± 0.9} & \textbf{68.9 ± 1.4}  & \textbf{16.7 ± 0.9} & \textbf{22.9 ± 1.2} & \textbf{12.5 ± 0.8} \\
\bottomrule
\end{tabular}}
\end{table}
As presented in Table \ref{tab:diffculty_results}, \textit{our proposed GeoNav agent demonstrates superior overall performance compared to all baseline methods. The advantages of our approach are consistent across varying difficulties, showcasing its strong generalization.} 
The advantage is particularly pronounced on hard tasks, where GeoNav achieves an SR of 16.7\%, in contrast to the second-best method, TopV-Nav, which only reaches 2.1\%.
These notable advances manifest the effectiveness of the well-designed spatial representations and flexible reasoning mechanisms.
More insightful observations can be obtained from the results.
\begin{itemize}
    \item Both rule-based methods and deep learning-based methods perform poorly on the task, and cannot navigate successfully. The results show that the CityNav challenge is difficult for simple rule-based or indoor approaches.
    \item Despite featuring strong multi-modal abilities, NavGPT and Qwen-VL-Max fail to navigate to the landmark, even with geographic priors. It underscores the vast decision space in urban navigation and the naive MLLMs' lack of spatial understanding.
    \item While MGP performs well in NE and SPL, its SR remains below 10\% across all difficulty levels. This aligns with our claim that top-down semantic maps like MGP aid coarse-grained navigation, but the lack of delicate, local representation and spatial reasoning affects the final localization.
    \item STMR and TopV-Nav w/ MNS perform comparably to MGP, highlighting the effectiveness of stage scheduling. Moreover, the comparison reveals that visual prompts (TopV-Nav) facilitate spatial reasoning more than metric maps (STMR).
\end{itemize}

\begin{table}[t]
    \centering
	\caption{The module ablation study on Validation Seen.}
    \resizebox{.65\linewidth}{!}{%
		\begin{tabular}{c|ccc|cccc}
			\hline
			\scriptsize \shortstack{\#} &
			{\footnotesize \textbf{SCM}}
                & \footnotesize \textbf{HSG}
			& \footnotesize \textbf{MNS}
			& {NE/m $\downarrow$}
			& {SR/\% $\uparrow$}
			& {OSR/\% $\uparrow$}
			& {SPL/\% $\uparrow$}
			\\
			\midrule
                \scriptsize \texttt{1}
                & \Checkmark&
                & 
                & 73.9
                & 15.5
                & 30.0
                 & 7.4
			\\
			\scriptsize \texttt{2}
                & \Checkmark & \Checkmark
                & 
                & 55.10
                & 17.78
                & 57.78
                & 10.43
			\\
                \scriptsize \texttt{3}
			    & \Checkmark &
                & \Checkmark
                & 55.74
                & 17.78
                & 44.44
                & 10.07
			\\
                \scriptsize \texttt{4}
			    & 
                & \Checkmark & \Checkmark
                & \textbf{53.47}
                & 8.89
                & 55.56
                & 5.46
			\\
                \scriptsize \texttt{5}
			    & \Checkmark & \Checkmark
                & \Checkmark
                & 59.86
                & \textbf{26.53}
                & \textbf{73.47}
                & \textbf{12.05}
			\\
			\bottomrule
		\end{tabular}
        }
	\label{tab:ablation_studies}
\end{table}
\subsection{Ablation Studies}
In this section, we first perform ablation tests on the three modules employed in our method. Subsequently, we evaluate the navigation accuracy at different stages of the three-stage navigation approach.
\subsubsection{Module Ablation}
\textit{Table \ref{tab:ablation_studies} presents the results of our ablation study, demonstrating that three modules—SCM, HSG, and MNS—are essential and work synergistically within GeoNav.} The contribution of each component is evident: removing the SCM (Method \#4) causes the success rate to collapse (SR -17.6\%), proving its critical role in global navigation, while adding either the HSG (Method \#2) or MNS (Method \#3) individually provides significant performance gains, highlighting their importance in local memory and strategic reasoning respectively.

\textit{Despite the highest performance on SR, OSR, and SPL, Method \#5 has a slightly higher average NE. We attribute this to a deliberate trade-off. The full method attempts more complex, longer-horizon plans, causing occasional error accumulation from the multi-stage strategy and more hallucinations of MLLMs within the abstract SCM.} This leads to a less precise final stopping point, indicating that our model prioritizes the primary goal of successful target identification over minimizing the final stopping distance.

\subsubsection{Evaluation on Goal Progress}
\begin{figure}
\centering
\includegraphics[width=.75\linewidth]{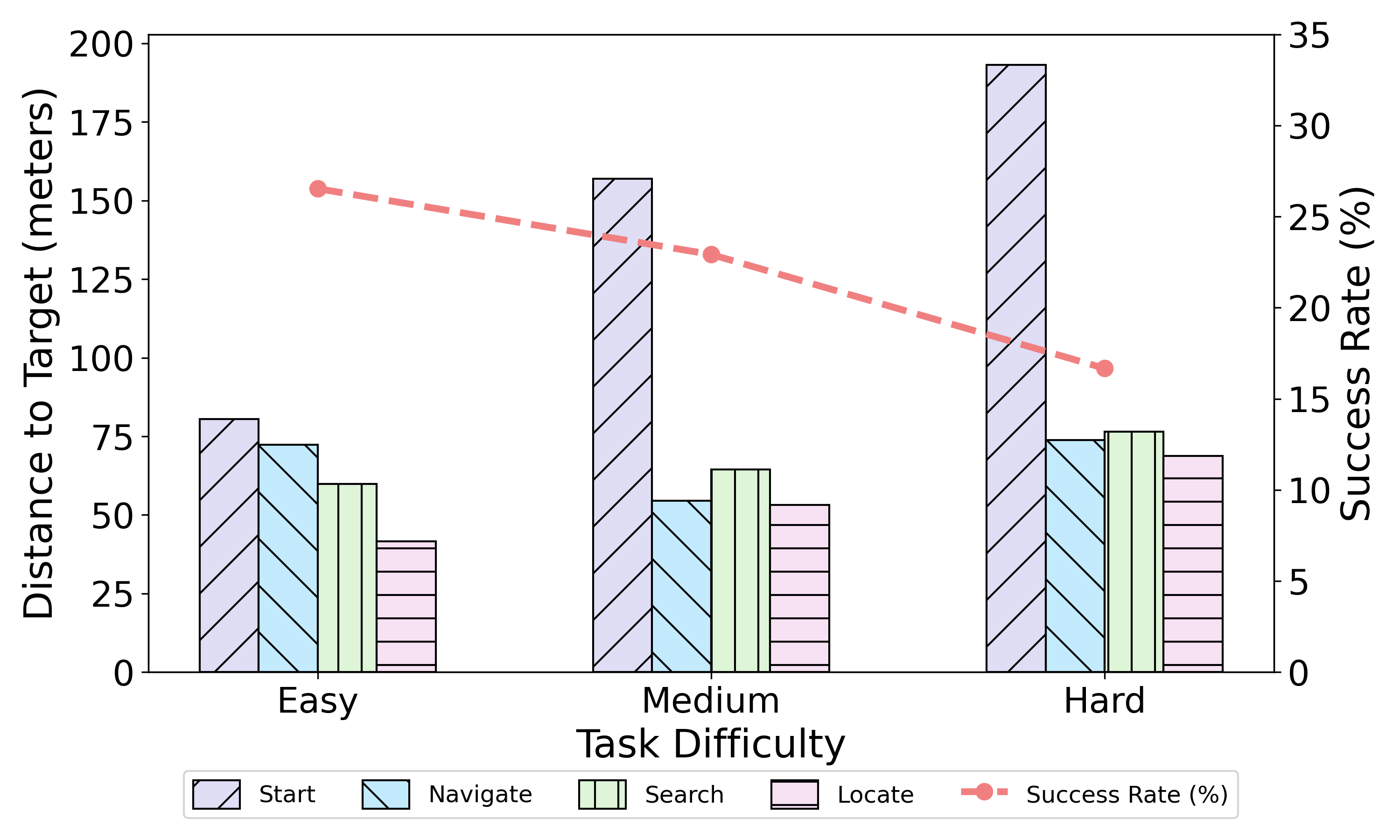}
\caption{Navigational progress in the three stages.}
\label{fig:exp_bar}
\end{figure}

In Sec. \ref{sec:intro}, we claimed that MNS gradually reduces the search space. Here, we examine whether the framework consistently leads the UAV to approach the target.
Some important observations can be obtained from Figure \ref{fig:exp_bar}.
\begin{itemize}
    \item Stage 1: The average distances to the targets are constantly reduced upon completing landmark navigation (i.e., the first stage) by 10.2\%, 65.23\%, and 61.8\%, respectively. 
    Especially for medium and hard settings, the initial distances are 156.87 and 193.19 m, respectively. This demonstrates the effectiveness of the navigation stage, particularly for long-distance tasks.
    \item Stage 2: GeoNav performs novelty-driven exploration in the search stage, it actively search the landmark and target, yielding uncertain remaining distances.
    \item Stage 3: Eventually, GeoNav retrieves the target after querying the HSG, demonstrating the effectiveness for target localization.
\end{itemize}

\subsection{Sensitivity Analysis}
In this section, we first evaluate performance across MLLMs of different sizes, followed by an analysis of navigation performance at varying altitudes.
\subsubsection{Effect of Different Model Sizes}
We select three small-size models for comparison, all of which can be feasibly deployed on UAVs. 
GeoNav then invokes the LLM in a limited manner. Specifically, once every ten action steps, to maintain seamless, real-time motion control. This strategy limits the total LLM calls to a maximum of 20 within a 200-step episode.
The results in Table \ref{tab:opensource_performance} show a clear correlation between the model scale and the capability of the MLLM. Nevertheless, 7B-parameter models demonstrate the fundamental navigation ability. For example, LLaVA-OneVision-7B achieves an SR of 9.52\%, surpassing that of QwenVL-Max. 



\begin{table}[t]
    \centering
    \begin{minipage}[t]{0.48\textwidth}
        \centering
        \caption{Average performance of open source models on the Validation Seen.}
        \label{tab:opensource_performance}
        \resizebox{\linewidth}{!}{%
            \begin{tabular}{l|ccc}
            \hline
            \textbf{Models} & NE/m $\downarrow$ & SR/\% $\uparrow$ & SPL/\%$\uparrow$ \\
            \midrule
            w/ LLaVA-OneVision-7B  & 93.40 & 9.52 & 3.52 \\
            w/ Qwen2.5-VL-7B  & 94.00 & 7.14 & 1.92 \\
            w/ Qwen2.5-VL-32B  & 88.7 & 12.85 & 4.76 \\
            w/ GPT-4o (our) & \textbf{56.93} & \textbf{22.00} & \textbf{4.98} \\
            \hline
            \end{tabular}
        }
    \end{minipage}
    \hfill 
    \begin{minipage}[t]{0.48\textwidth}
        \centering
        \caption{Average performance comparison between varying observation altitudes.}
        \label{tab:altitude_performance}
        \resizebox{\linewidth}{!}{%
            \begin{tabular}{l|cccc}
            \hline
            \textbf{Altitude (z)} & GSD/m & NE/m & SR/\% & SPL/\%\\
            \midrule
            20.0m & $4.17 \times 10^{-2}$ & \textbf{49.41} & 16.67 & \textbf{13.64} \\
            50.0m & $1.04 \times 10^{-1}$ & 59.86 & \textbf{26.53}  & 10.43\\
            80.0m & $1.67 \times 10^{-1}$ & 99.48 & 7.14 & 1.68 \\
            100.0m& $2.08 \times 10^{-1}$  & 113.17 & 7.14 & 2.22\\
            \hline
            \end{tabular}
        }
    \end{minipage}
\end{table}

\subsubsection{Effect of Observation Altitudes}
Our analysis of observation altitude reveals a trade-off between field-of-view and ground resolution, which is measured by the ground distance represented by a pixel, i.e., ground sample distance (GSD). At a low altitude of 20.0m, the agent achieves maximum navigation accuracy, evidenced by the lowest NE of 49.4m and the highest SPL of 13.64\%. This suggests that a high-resolution view is optimal for final target identification and stopping maneuvers. However, the peak overall task completion was observed at 50.0m, indicating a balance between a wide perceptual field for spatial reasoning and sufficient detail for object recognition. We set it as the default in other experiments. Conversely, performance collapses at higher altitudes of 80.0m and 100.0m, where the SR plummets to 7.14\%. This decline directly correlates with poor image quality, as quantified by GSD, highlighting the limitations of vision models in interpreting low-resolution imagery for nuanced, language-guided tasks.

\begin{figure*}
\centering
\includegraphics[width=.95\textwidth]{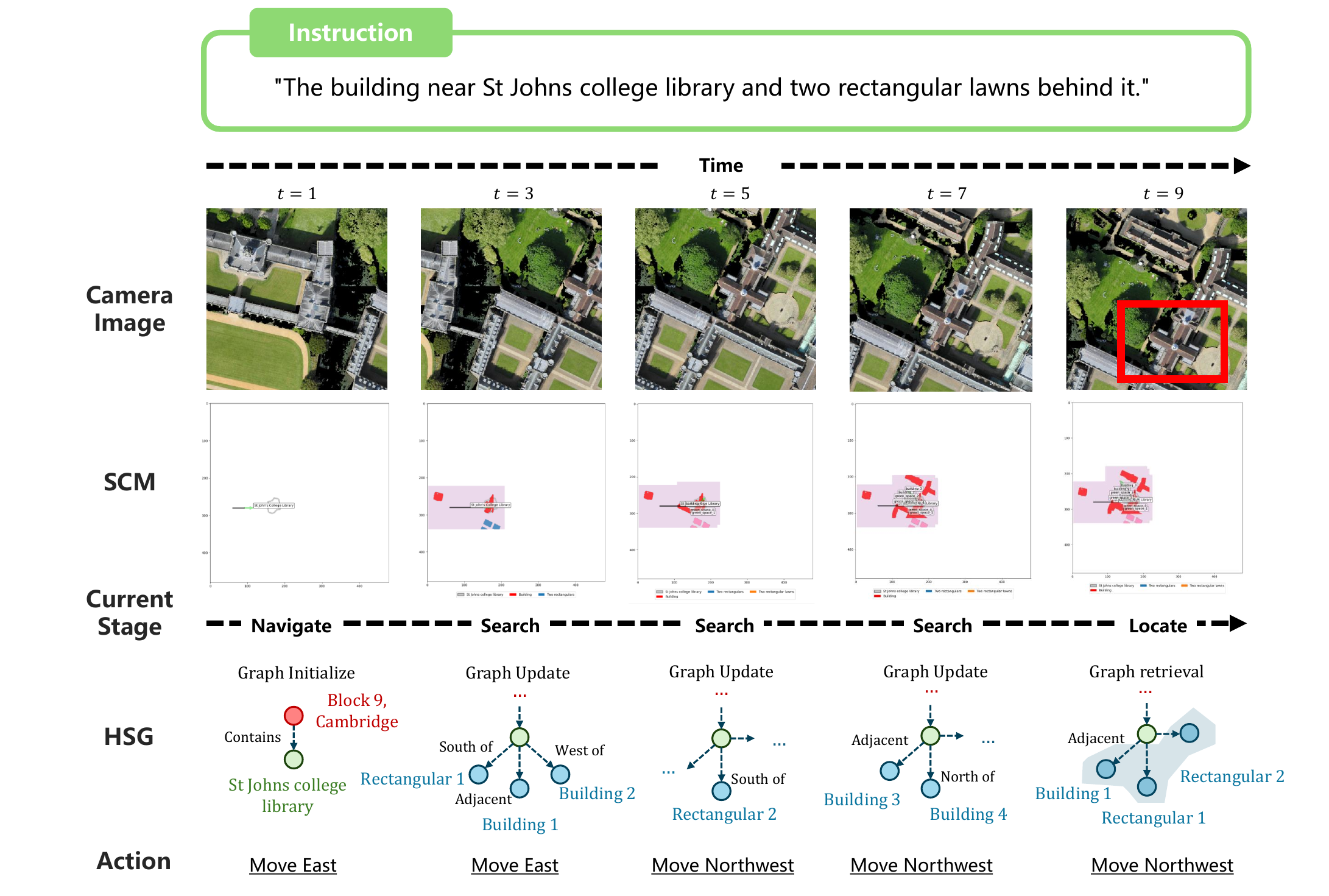}
\caption{Case 1 shows a qualitative example of our GeoNav agent performing the language-goal aerial navigation task on the CityNav benchmark.}
\label{fig:case}
\end{figure*}

\subsection{Qualitative Analysis}

In this subsection, we visualize the navigation process of GeoNav.
Figure \ref{fig:case} presents a successful case on the Validation Seen split. This example shows that in the beginning of search phase, HSG only includes block and landmark nodes. GeoNav processes the SCM obtained after processing the camera image and infers the forward movement. Then, after arriving near St Johns college library, the drone starts searching based on the SCM. At the same time, HSG starts updating the object node at this time. Finally, when the exploration coverage condition is met, it switches to the localization phase. GeoNav obtains the shadow-covered subgraph through multi-step queries of complex target descriptions, and infers the node that is most likely to be the target. For more examples, please refer to \ref{appendix_cases}.

\subsection{Analysis of Computational Complexity}
GeoNav maintains quadratic-time computational complexity owing to its hierarchical structured memory. The SCM and HSG are updated with linear and near-quadratic time complexity with respect to the object number, while each update is constrained within a limited context, ensuring practical efficiency. More importantly, the MLLM is invoked at the stage level, rather than at each action step. This strategy decouples control from reasoning, thereby reducing both token consumption and inference latency. As a result, GeoNav achieves efficient and interpretable navigation with minimal frequent invocations, making it deployable for real-time use on aerial platforms.

\section{Discussion and Conclusions}
This work introduces GeoNav, an agentic MLLM framework that tackles complex language-guided aerial navigation by mimicking human coarse-to-fine reasoning. At its core is a novel dual-memory system designed to empower the MLLM: a top-down cognitive map provides global, landmark-based cues, while a hierarchical scene graph (HSG) manages dynamic, local object relationships for precise target localization.

On the challenging, long-range CityNav benchmark, GeoNav sets a new state-of-the-art, decisively outperforming existing methods in both success rate and navigation precision. Our ablation studies validate this architecture, confirming that the complete three-stage planning process is essential for task completion, while the dual-memory modules are critical for accurate navigation and positioning. Furthermore, our design achieves a practical balance between performance and efficiency. By strategically invoking the MLLM only for high-level, critical decisions, it mitigates the high token consumption and latency typically associated with large models.
This philosophy is exemplified by phase-switching design. The transition relies on a simple mechanism to counteract the decision limitations of current MLLMs, which proved effective in practice. The coarse-to-fine pipeline naturally separates global movement from local exploration without additional task-specific tuning.

However, a performance and efficiency gap to humans persists, charting a clear path for future endeavors. Key directions to bridge this gap include: (1) exploring end-to-end models that unify spatial memory and inference to reduce token consumption radically; (2) enabling adaptive altitude adjustment to grant the agent true 3D spatial reasoning.

\appendix
\label{app1}

\section{Stage Switch Condition}
\label{appendix_stage}

Our task planning module decomposes the complex navigation task into an executable sequence of sub-goals. This process utilizes three strategies, and the switch conditions are defined based on the agent's pose $\xi_t$, visual observation $I_t$, and the provided geographic priors $\mathcal{K}$.

\textbf{1. Landmark Navigation}
The goal of this stage is to use the geographic priors $\mathcal{K}$ (map) to navigate toward a specified landmark. Given the agent's current position $P_c$ (from $\xi_t$) and the target landmark's position $P_t$ (from $c_i$ in $\mathcal{K}$):
\begin{align}
    P_c &= (x, y, z) \\
    P_t &= (x', y', z')
\end{align}
The agent calculates the Euclidean distance to the target as $d = \|P_t - P_c\|$. This stage terminates and switches to the next strategy when the agent is sufficiently close to the landmark, defined by the condition $d \le d_{\text{nav\_thresh}}$, i.e. 50m. It is worth noting that this stage is not required to pinpoint the exact target. Instead, it serves to reliably transport the agent into the 'actionable range' of the subsequent Search stage, where fine-grained positioning is handled via active visual exploration.

\textbf{2. Visual Search}
The objective of this stage is to use the egocentric top-down RGB image $I_t$ to visually confirm the landmark or search for the specific target described in the language instruction $\mathcal{T}$. This strategy is critical for aligning the abstract geographic priors $\mathcal{K}$ (map) with the real-time visual observation $I_t$. 

\textbf{3. Final Localization}
After the target object $P_o$ has been confirmed (combining information from $\mathcal{K}$, $\mathcal{T}$, and $I_t$), this stage is activated. The system uses the confirmed target coordinates $P_o = (x_o, y_o, z_o)$. The agent navigates directly toward $P_o$. The episode concludes successfully when the agent executes the `Stop` action within the success radius (defined by our setting as 20 meters), i.e., $\|P_o - P_c\| \le 20\text{m}$.

\begin{tcolorbox}[
  title=\textbf{1. An Example of the Task Decomposition},
  label={box:task},
  colback=red!5,
  colframe=red!75!black,
  fonttitle=\bfseries,
  boxrule=0.5pt,
  left=2pt,right=2pt,top=2pt,bottom=2pt,
  before skip=4pt,after skip=4pt
]
\setlength{\baselineskip}{0.8\baselineskip} 

\textbf{Instruction:} Find the dark red car in the middle row of the West Midlands Police Custody Suite parking lot, between a red car and a gray car.

\textbf{Plan:} Locate the dark red car situated in the middle row of the custody suite parking lot. The UAV should navigate toward the suite, scan the specified area, and identify the target vehicle between the red and gray cars.

\textbf{Sub-goals:}
\begin{enumerate}
  \item \textbf{Goal:} Navigate near the West Midlands Police Custody Suite. 
        \textbf{Desired State:} UAV is positioned 10–20 m southeast of the suite. 
        \textbf{Strategy:} Navigate.
  \item \textbf{Goal:} Search the middle row of the parking lot. 
        \textbf{Desired State:} UAV scans the entire row, identifying all vehicles. 
        \textbf{Strategy:} Search.
  \item \textbf{Goal:} Locate the dark red car between a red and a gray car. 
        \textbf{Desired State:} Target car identified and location confirmed. 
        \textbf{Strategy:} Locate.
\end{enumerate}
\end{tcolorbox}

\section{Multi-modal Chain of Thought}\label{appendix_cot}
\begin{figure}[htbp]
\centering
\begin{subfigure}[t]{0.49\textwidth}
    \centering
    \includegraphics[width=\linewidth]{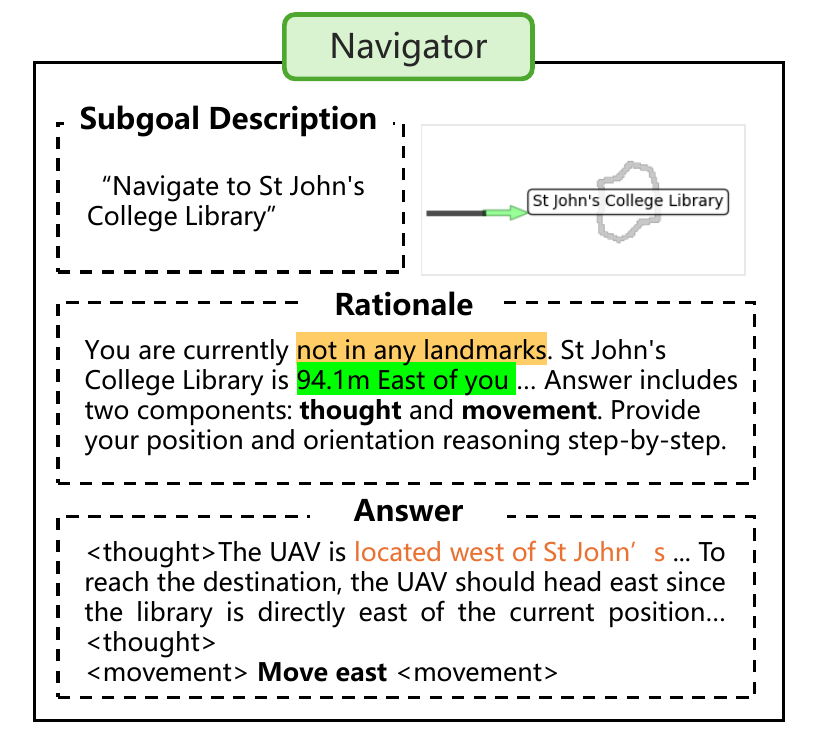}
    \caption{Prompting for landmark navigation}
    \label{fig:cot_1}
\end{subfigure}
\hfill
\begin{subfigure}[t]{0.49\textwidth}
    \centering
    \includegraphics[width=\linewidth]{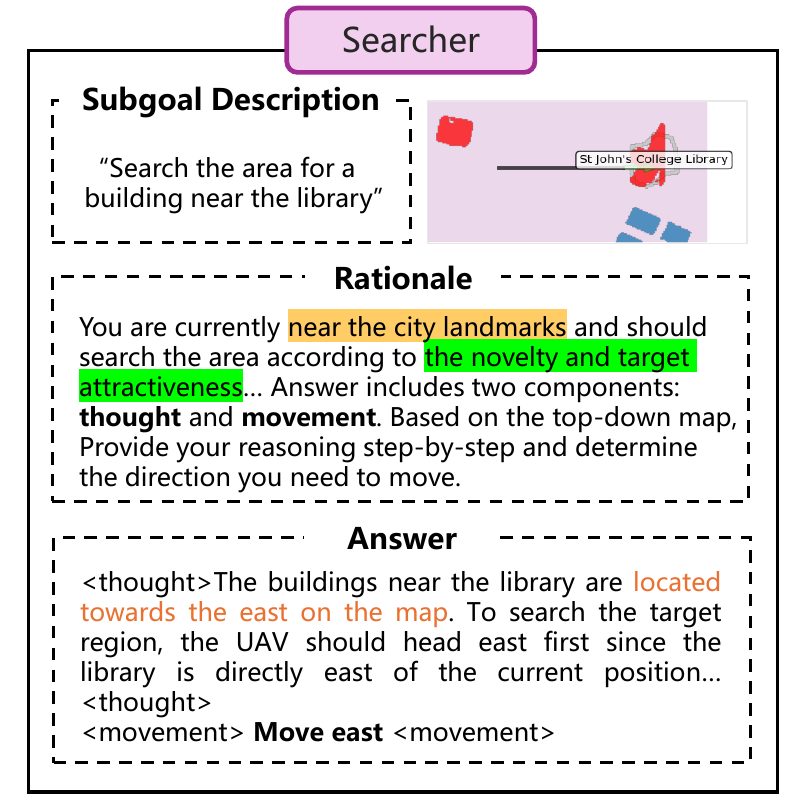}
    \caption{Prompting for target search}
    \label{fig:cot_2}
\end{subfigure}

\vspace{6pt} 


\caption{The multi-modal reasoning of MLLM in GeoNav.}
\label{fig:cot}
\end{figure}

Figure \ref{fig:cot} details our Chain-of-Thought (CoT) prompting strategy. For instance, in Figure \ref{fig:cot} (a), GeoNav first generates a stage-aware subgoal to specify the next landmark, then uses the top-down cognitive map with the navigational rationale to constrain the response of the MLLM.

\section{Schematic Cognition Map Generation details}\label{appendix_SCM}

We detail the fusion of geographic priors and visual data into the cognition map, explaining how visual foundation models are utilized to update the spatial memory $M_t = M_{t-1} \cup \Delta_{\text{perceive}}(I_t)$.

\subsection{Geographic Priors}
Prior landmark contours are retrieved from OpenGIS data and converted to pixel coordinates $\text{(col, row)}$ given map boundaries $(x_{\min}, y_{\min}, x_{\max}, y_{\max})$ and a scale $s$ (pixels/meter):
\begin{equation}
    \text{(col, row)} = \big( \operatorname{round}((x - x_{\min})s), \operatorname{round}((y_{\max} - y)s) \big).
\end{equation}
We annotate these priors with landmark labels and overlay the UAV's trajectory (position, heading) to ground the MLLM in the global context.

\subsection{Visual Observations and Fusion}
GeoNav pipeline includes perception via foundation models, spatial projection and map update. First, we employ \textbf{Grounding DINO} for open-vocabulary object detection to obtain bounding boxes. The boxes prompt \textbf{SAM} to generate pixel-wise segmentation masks for the detected objects. Second, the extracted masks are projected onto the world map frame. We apply a rotation based on the UAV's yaw $\theta$ and translation by its position $(x_p, y_p)$ to obtain the world coordinates $(x_w, y_w)$. This transformation is defined as:
\begin{equation}
\begin{pmatrix} x_w \\ y_w \end{pmatrix}
=
-\begin{bmatrix}
\cos\theta & -\sin\theta\\
\sin\theta & \cos\theta
\end{bmatrix}
\begin{pmatrix} x_c \\ y_c \end{pmatrix}
+
\begin{pmatrix} x_p \\ y_p \end{pmatrix}.
\end{equation}

Third, we transform semantic masks into world regions. These projected regions constitute $\Delta_{\text{perceive}}(I_t)$, which are then spatially registered and merged into the updated map $M_t$.

\subsection{Graph Generation}\label{appendix_HSG}
\begin{figure}[h!]
    \centering
    \includegraphics[width=.8\linewidth]{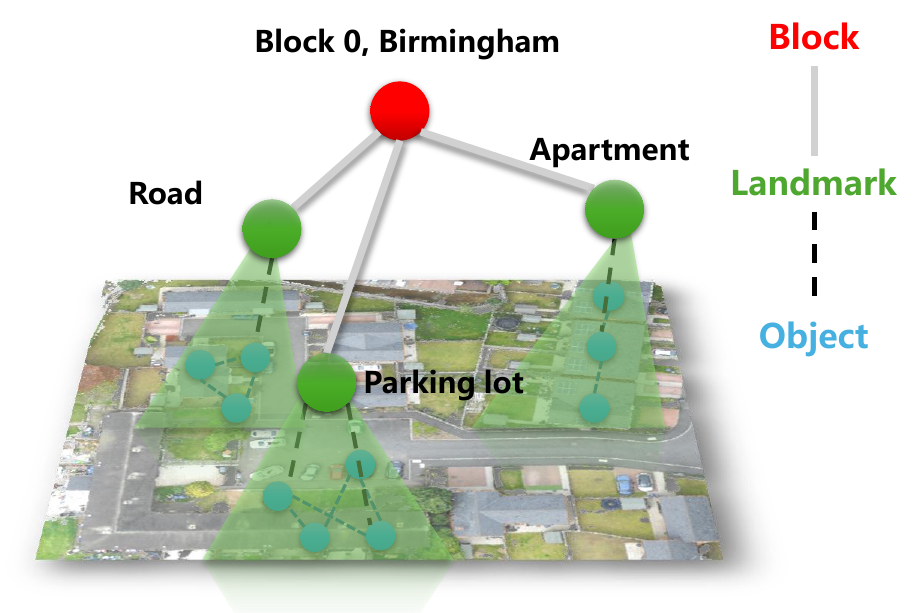}
    \caption{The Visualization of Hierarchical Scene Graph.}
    \label{fig:sg_app}
\end{figure}

In this subsection, we present the details of constructing node and edge relationships, the query templates for MLLM complex statements.
As shown in the text box, we determine node requirements by specifying the object type label and possible attributes. The nodes employ R-tree structures (Figure \ref{fig:sg_app}) for efficient spatial storage and reasoning, supporting fast proximity, containment, and landmark queries to detect duplicates and locate objects within landmark regions.
For the edge requirement, we restrict the spatial semantics of edges by defining topological relationships and absolute orientations.

\begin{tcolorbox}[
  title=\textbf{Node and Edge Requirements},
  colback=blue!5,
  colframe=blue!75!black,
  fonttitle=\bfseries,
  boxrule=0.5pt,
  left=2pt,right=2pt,top=2pt,bottom=2pt,
  before skip=4pt,after skip=4pt
]\label{box:node_edge}
\setlength{\baselineskip}{0.85\baselineskip}

\textbf{Node Requirements.}
Each node has a unique \texttt{id}. Mandatory attributes: 
\textbf{\texttt{object\_type}} $\in$ \{``vehicle'', ``road'', ``building'', ``parking\_lot'', ``green\_space'', …\}, 
and \textbf{\texttt{bbox}} = [xmin, ymin, xmax, ymax].  
Optional (if observable): \textbf{\texttt{color}} $\in$ \{``white'', ``black'', ``red'', ``gray'', ``blue'', ``green'', ``brown'', ``silver''\}.

\textbf{Edge Requirements.}
Only the following relation labels are allowed.  
\textit{Topological:} \textbf{\texttt{contains}} (A in B), \textbf{\texttt{adjacent\_to}} (side-by-side), \textbf{\texttt{near\_corner}} (near corner).  
\textit{Directional (absolute, aerial view):}  
Primary — \textbf{\texttt{north\_of}}, \textbf{\texttt{south\_of}}, \textbf{\texttt{east\_of}}, \textbf{\texttt{west\_of}};  
Diagonal — \textbf{\texttt{northeast\_of}}, \textbf{\texttt{northwest\_of}}, \textbf{\texttt{southeast\_of}} ...
\end{tcolorbox}
\section{Implementation Details for Hierarchical Scene Graph}
\label{appendix:hsg_details}
The chain filters nodes that satisfy the conditions in the operation sequence. If the initial query is unsuccessful, the system employs recursive techniques to refine the search criteria, gradually narrowing the focus until the candidate nodes are found.

\begin{tcolorbox}[
  title=\textbf{Prompt for Query Operation Chain},
  label={box:query_chain},
  colback=green!5,
  colframe=green!75!black,
  fonttitle=\bfseries,
  boxrule=0.5pt,
  left=2pt,right=2pt,top=2pt,bottom=2pt,
  before skip=4pt,after skip=4pt
]
\setlength{\baselineskip}{0.8\baselineskip} 
\small 

\textbf{QUERY\_OPERATION\_CHAIN\_PROMPT = \texttt{"""}}  
\{instruction\}. Convert instructions into query operations.

\textbf{\texttt{get\_geonode\_by\_name(name),}}\textbf{\texttt{get\_child\_nodes(parent, rel),}} \textbf{\texttt{filter\_by\_class(cls)...}}

\textbf{Notes:}  
``In front of'' → \texttt{north\_of};  
``Behind'' → \texttt{south\_of};  
``On the road'' → \texttt{contains}.

\vspace{0.4em}
\noindent\textbf{Example:} ``A white car parked on Davey Road with one gray car in front of it.''  
{\footnotesize
\begin{verbatim}
[
 {"method":"get_geonode_by_name","args":["Davey Road"]},
 {"method":"get_child_nodes","kwargs":{"relation_type":"contains"}},
 {"method":"filter_by_class","args":["vehicle"]},
 {"method":"filter_by_attribute","args":["color","white"]}
]
\end{verbatim}
}
\texttt{"""}
\end{tcolorbox}
\subsection{Incremental Graph Update and Hyperparameters}
\label{appendix:implementation}
As detailed in the main manuscript, the HSG is updated incrementally to prevent the creation of duplicate nodes from overlapping camera frames. This section provides the precise implementation details for the node-merging mechanism and lists all relevant hyperparameters for reproducibility.

\paragraph{Node Merging Criterion}
To decide whether a new node candidate should be merged with an existing node, we compute a comprehensive similarity score $S_{ij}$ that fuses both spatial and semantic information:
\begin{equation}
\label{eq:appendix_similarity}
S_{ij} = (1-\gamma) \, S_{ij}^{\text{spatial}} + \gamma \, S_{ij}^{\text{semantic}},
\end{equation}
where both components are formulated as bounded similarity scores. The spatial similarity $S_{ij}^{\text{spatial}}$ converts the Euclidean distance between two nodes into a score using a Gaussian kernel:
\begin{equation}
\label{eq:appendix_spatial_similarity}
S_{ij}^{\text{spatial}} = \exp(-\|p_i - p_j\|_2^2 / \sigma^2),
\end{equation}
The semantic similarity $S_{ij}^{\text{semantic}}$ is the cosine similarity between the visual feature embeddings of the two nodes. If the combined score $S_{ij}$ exceeds the threshold $\rho$, the nodes are merged.
\paragraph{Chain of Query Operations}
In query operations, an instruction is decompose into a structured chain of operations (textbox \ref{box:query_chain}). The JSON defines query steps, each specifying a method (e.g., identify node by name) and parameters. Simple queries are handled directly, while complex instructions are decomposed into multiple subqueries.

\begin{figure*}
\centering
\includegraphics[width=.95\textwidth]{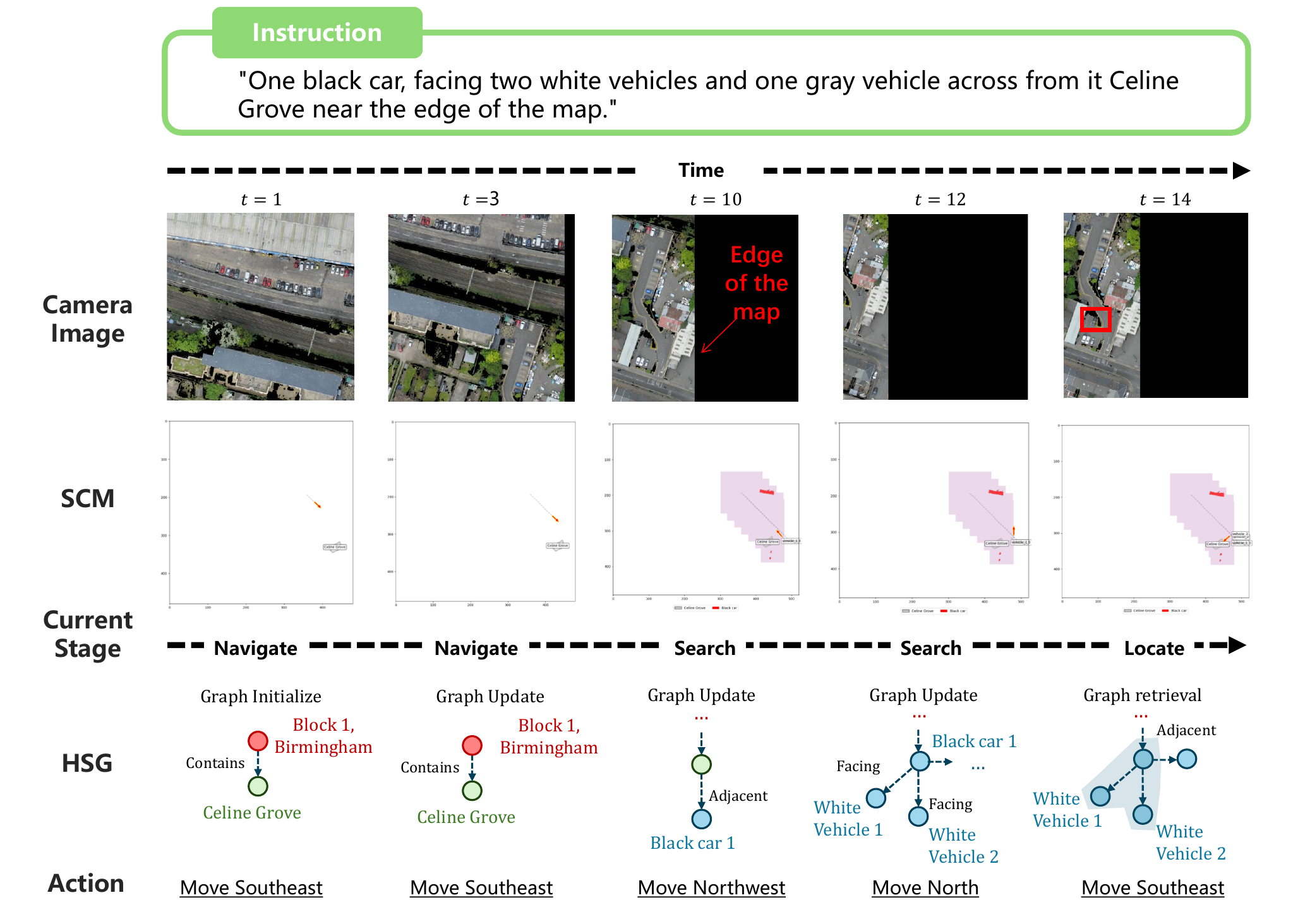}
\caption{Case 2 exhibits a qualitative example of our GeoNav agent performing 
LLM-Driven Target Retrieval for the language-goal aerial navigation task.}
\label{fig:case2}
\end{figure*}
\section{Implementation Details}\label{appendix_implementation}
\noindent\textbf{Our code is available at:}
\href{https://anonymous.4open.science/r/GeoNavAgent-42D8}{\texttt{Code}} \;|\;
\href{https://figshare.com/s/24c633980aadb14e397b}{\texttt{Example Dataset}}.
Hyperparameters are as follows: the similarity weight $\gamma=0.95$, the merging threshold $\rho=0.8$. The spatial scaling factor $\sigma=10.0$m, aligned with the typical urban scale for spatial similarity. The altitude is fixed at $50$m, the detector confidence threshold is $0.20$; and the maximum timesteps are $20$, including up to $10$ steps for \textit{Navigate} and $6$ steps for \textit{Search} phases. Our latency profiling (including perception, memory updates, and GPT-4o inference) indicates that the MLLM reasoning is fully masked by the UAV flight window for a 10-step (50 m) segment, supporting the invoking of MLLMs once every 10 steps.

\section{Case Study}
\label{appendix_cases}
\subsection{Good Cases}
Figure \ref{fig:case2} shows a successful case under the Hard difficulty level on the \textit{Val\_seen} split. During the construction of the HSG, incorrect edge attributes led to failures in the chained query operations. However, as illustrated in the text box \ref{box:trace1}, GeoNav successfully retrieved the most probable node through its robust fallback mechanism and navigated to the correct destination.

\begin{tcolorbox}[
  colback=gray!5!white,
  colframe=gray!75!black,
  title=Query Trace (Case 2),
  label={box:trace1},
  boxrule=0.4pt,
  left=2pt,right=2pt,top=2pt,bottom=2pt,
  before skip=4pt,after skip=4pt
]
\setlength{\baselineskip}{0.8\baselineskip}
\small
\texttt{get\_geonode\_by\_name} → 1 node. 
\texttt{get\_child\_nodes} → 0 → fallback: 1 node. 
\texttt{filter\_by\_class} → fallback: 1 node. 
\texttt{filter\_by\_attribute} → fallback: 1 node. 
Query completed: 1 node found — \texttt{TrafficRoad\_Celine Grove (geo)}, distance = 18.87\,m. 
\textbf{Target reached.}
\end{tcolorbox}



\subsection{Failure Cases}
\begin{figure*}
\centering
\includegraphics[width=.95\textwidth]{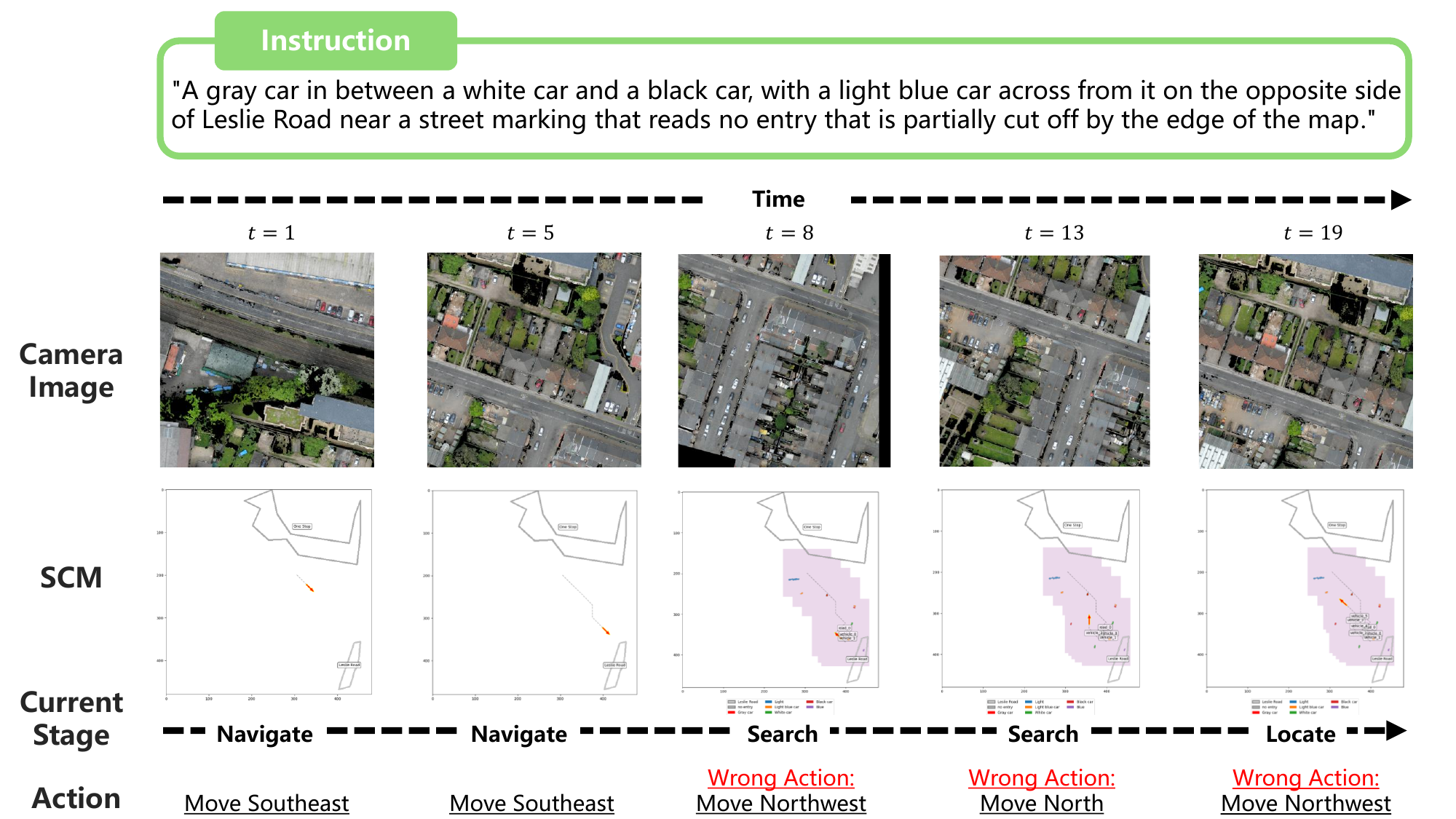}
\caption{Case 4 is a failure case of our GeoNav agent in extreme complex instruction and hard level navigation task.}
\label{fig:case4}
\end{figure*}

Complex navigation instructions and long flight trajectories can lead to errors.
Figure \ref{fig:case4} illustrates a failure case in the hard task difficulty where the MLLM produces hallucinated actions during the search phase, leading the agent to deviate from the correct exploration direction. At timestep 19, the UAV revisit the position at timestep 5. As the distance increases, the correct target remains unexplored, ultimately resulting in task failure.

As shown in the text box, case 5 presents a failure case of the HSG. Specifically, due to the high similarity between neighboring vehicle nodes, some nodes were incorrectly removed during the deduplication process. As a result, the query instruction failed to retrieve a matching node.

\begin{tcolorbox}[
  colback=gray!5!white,
  colframe=gray!75!black,
  title=Query Trace (Case 5),
  label={box:trace2},
  boxrule=0.4pt,
  left=2pt,right=2pt,top=2pt,bottom=2pt,
  before skip=4pt,after skip=4pt
]
\setlength{\baselineskip}{0.8\baselineskip}
\small
\texttt{get\_geonode\_by\_name} → 1 node; 
\texttt{get\_child\_nodes} → fallback: 1 node; 
\texttt{filter\_by\_class} → fallback: 1 node. 
Query completed: 1 node found — \texttt{TrafficRoad\_Bragg Road (geo)}, distance = 100.41\,m.
\end{tcolorbox}


 \bibliographystyle{elsarticle-num} 
 \bibliography{cas-refs}






\end{document}